\documentclass[10pt,twocolumn,letterpaper]{article}

\usepackage[pagenumbers]{iccv} 

\usepackage[utf8]{inputenc}         %
\usepackage[T1]{fontenc}            %
\usepackage{url}                    %
\usepackage{booktabs}               %
\usepackage{amsfonts}               %
\usepackage{nicefrac}               %
\usepackage{microtype}              %
\usepackage[dvipsnames]{xcolor}     %

\usepackage[utf8]{inputenc}         %
\usepackage[T1]{fontenc}            %
\usepackage{url}                    %
\usepackage{booktabs}               %
\usepackage{amsfonts}               %
\usepackage{nicefrac}               %
\usepackage{microtype}              %
\usepackage{xcolor}                 %

\usepackage[flushleft]{threeparttable}
\usepackage{float}
\usepackage{graphicx}
\usepackage{multirow}
\usepackage{xspace}
\usepackage{natbib}
\usepackage{enumitem}
\usepackage[font=small]{caption}
\usepackage{autobreak}
\usepackage{sidecap}
\usepackage{wrapfig}
\usepackage[toc, page, header]{appendix}
\usepackage{makecell}

\newcommand{\benchopt}{OHRBench\xspace}

\usepackage{tabularx}
\usepackage{soul, color, xcolor}
\usepackage{adjustbox}
\usepackage{colortbl}
\usepackage{amsmath,amsthm,amssymb}

\newenvironment{talign*}
{\csname align*\endcsname}
{\endalign}

\definecolor{iccvblue}{rgb}{0.21,0.49,0.74}
\usepackage[pagebackref,breaklinks,colorlinks,allcolors=iccvblue]{hyperref}
\usepackage[accsupp]{axessibility}  
\usepackage{float}
\usepackage{placeins}
\usepackage{needspace} 

\def\paperID{3742} 
\def\confName{ICCV}
\def\confYear{2025}

\title{OCR Hinders RAG: Evaluating the Cascading Impact of OCR on Retrieval-Augmented Generation}

\author{
Junyuan Zhang$^{1,3*}$\quad
Qintong Zhang$^{1,2*}$\quad
Bin Wang$^{1*}$\quad
Linke Ouyang$^{1}$\quad
Zichen Wen$^{1,4}$\quad\\
Ying Li$^{5}$\quad
Ka-Ho Chow$^{3}$\quad
Conghui He$^{1\ddag}$\quad
Wentao Zhang$^{2}$\\
$^1$Shanghai AI Laboratory\quad
$^2$Peking University\quad
$^3$The University of HongKong\quad\\
$^4$Shanghai Jiaotong University\quad
$^5$Beihang University
}

\begin{document}
\maketitle
\renewcommand{\thefootnote}{}
\footnotetext{$^*$ These authors contributed equally to this work.}
\footnotetext{$^\ddag$ Corresponding author (heconghui@pjlab.org.cn).}
\renewcommand{\thefootnote}{\arabic{footnote}}
\begin{abstract}
Retrieval-augmented Generation (RAG) enhances Large Language Models (LLMs) by integrating external knowledge to reduce hallucinations and incorporate up-to-date information without retraining.
As an essential part of RAG, external knowledge bases are commonly built by extracting structured data from unstructured PDF documents using Optical Character Recognition (OCR).
However, given the imperfect prediction of OCR and the inherent non-uniform representation of structured data, knowledge bases inevitably contain various OCR noises.
In this paper, we introduce \benchopt, the first benchmark for understanding the cascading impact of OCR on RAG systems.
\benchopt includes 8,561 carefully selected unstructured document images from seven real-world RAG application domains, along with 8,498 Q\&A pairs derived from multimodal elements in documents, challenging existing OCR solutions used for RAG.
To better understand OCR's impact on RAG systems, we identify two primary types of OCR noise: \textit{Semantic Noise} and \textit{Formatting Noise} and apply perturbation to generate a set of structured data with varying degrees of each OCR noise.
Using \benchopt, we first conduct a comprehensive evaluation of current OCR solutions and reveal that none is competent for constructing high-quality knowledge bases for RAG systems.
We then systematically evaluate the impact of these two noise types and demonstrate the trend relationship between the degree of OCR noise and RAG performance.
Our \benchopt, including PDF documents, Q\&As, and the ground truth structured data are released at: \url{https://github.com/opendatalab/OHR-Bench}
\end{abstract}
\section{Introduction}
\label{sec:intro}

Retrieval Augmented Generation (RAG) enhances Large Language Models (LLMs) by integrating external knowledge~\cite{gao2023retrieval,lewis2020retrieval}, enabling them to respond accurately to queries beyond their training corpus, such as recent news or proprietary content, and reducing hallucinations~\cite{lewis2020retrieval,izacard2023atlas,ram2023context,team2025novelseek}.
This is achieved through a retrieval-then-grounding approach, where relevant documents are retrieved from external knowledge bases and incorporated into the LLM's prompt for grounding.

As an essential component of RAG systems, the knowledge base defines the scope and quality of documents that RAG can access.
Given that a vast amount of real-world knowledge resides in unstructured documents, such as scanned PDFs, constructing an external knowledge base often relies on Optical Character Recognition (OCR)~\footnote{We employ the General OCR concept for document parsing from GOT-OCR2.0~\cite{wei2024general}, which includes, text recognition, multimodal data extraction (table, formula, and chart recognition), and reading order restoration.} to parse structured data from these unstructured PDF documents~\cite{zhang2024document,hui2024uda}.
For instance, MinerU~\cite{wang2024mineru} takes raw PDFs as input and extracts plain text, formulas, and tables into structured formats for subsequent RAG applications.
However, imperfect predictions of OCR and non-uniform representations of parsing results impair the construction of a high-quality knowledge base for RAG.
To be specific, despite advancements in OCR~\cite{wang2024mineru,wei2024general,blecher2023nougat}, even the leading model cannot achieve perfect accuracy across all scenarios~\cite{liu2024ocrbenchhiddenmysteryocr,zhang2024document}.
Furthermore, structural data like table can inherently be parsed in different representation, such as Markdown or LaTeX.
These issues introduce OCR noise in parsing results and diminish the quality of the knowledge base.
Considering RAG is sensitive to input noise~\cite{fang2024enhancing,xu2024unsupervised,cho2024typos}, recent works race on downstream RAG components, including more precise retrievers~\cite{bge-m3,moreira2024nv,li2023towards} and more advanced LLMs~\cite{achiam2023gpt,dubey2024llama,qwen2,fang2024enhancing}.
However, the quality of OCR-based external knowledge bases and its cascading impact on these downstream RAG components have received less attention, which highlights a critical but unaddressed gap: \textit{the absence of benchmarks to assess OCR's cascading impact on each component and entire system of RAG.}

\begin{figure*}[t]
    \centering
    \includegraphics[width=0.98\linewidth]{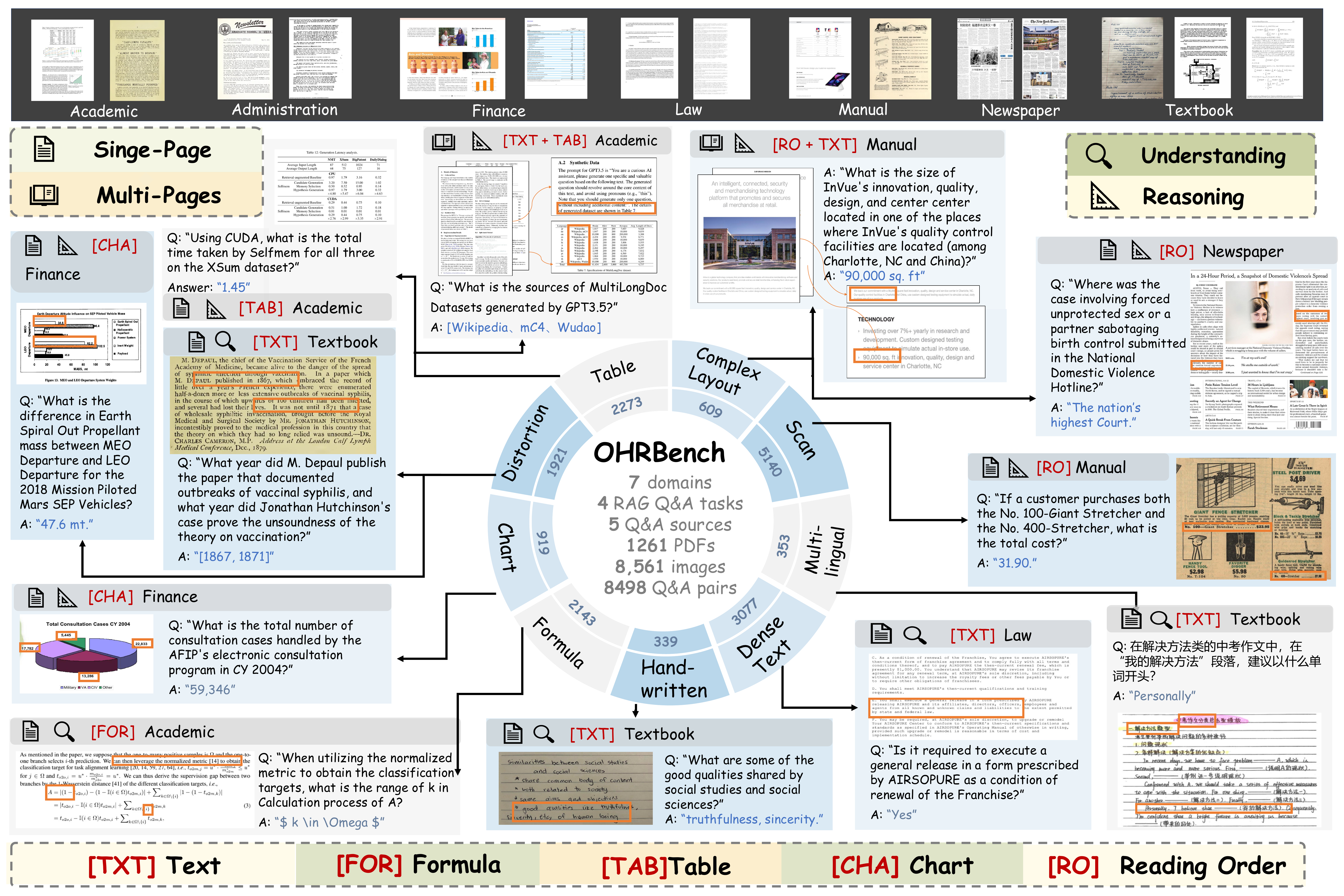}
    \vspace{-4mm}
    \caption{Our \benchopt comprises documents from 7 domains, 9 challenging attributes for OCR, 4 types of Q\&A tasks, and 5 Q\&A evidence sources. 
    Each number indicates the count of PDF pages with that attribute. 
    Criteria for these attributes can be found in Appendix~\cref{appendix:complex_layout}}
    \label{fig:overview}
    \vspace{-6mm}
\end{figure*}


Existing benchmarks either evaluate RAG holistically without fine-grained assessment~\cite{zou2024docbench}, consider limited OCR solutions without accounting for the noise they introduce~\cite{hui2024uda,faysse2024colpali}. Additionally, they lack documents that present more diverse OCR challenges, such as scanned historical, multilingual, and handwritten documents.
To fill this gap, we introduce \benchopt, a question-answering benchmark designed to evaluate OCR's cascading impact on each component and entire systems of RAG in two ways.
First, we construct a document-based RAG Q\&A dataset comprising complex, unstructured PDF documents from 7 RAG real-world application areas: \textit{Textbook, Law, Finance, Newspaper, Manual, Academic and Administration}.
As detailed in~\cref{tab:dataset_statistics} and~\cref{fig:overview}, we have collected 8,561 document images featuring attributes that challenges the creation of high-quality knowledge bases for RAG systems.
We also provide diverse Q\&A pairs which not only span realistic RAG tasks, including understanding, reasoning, and multi-page questions, but also features evidence sourced from key components of OCR in document parsing, making them ideal for assessing the OCR's impact on RAG performance.
Second, we identify two primary OCR noise types: \textit{Semantic Noise}, resulting from prediction errors, and \textit{Formatting Noise}, arising from diverse document element representation.
By systematically introducing these noise types into documents, we generate perturbed structured data with varying degrees of noise, enabling further exploration of the quantitative relationship between OCR noise and RAG performance.

With \benchopt, we first conduct a comprehensive benchmark on current OCR solutions, including pipeline-based OCR systems~\cite{wang2024mineru,marker}, end-to-end OCR models~\cite{blecher2023nougat,wei2024general} and Vision-Language Models (VLMs) for OCR~\cite{bai2025qwen2,wang2024qwen2,chen2024far,chen2024expanding,wen2025stop,wen2025token}. 
We reveal that even the best OCR solutions exhibit a performance gap of 14\% at least, compared to the ground truth structured data, facilitating the importance of mitigating OCR noise in RAG systems.
Further experiments on different types of OCR noise uncover that \textit{Semantic Noise} consistently exert a significant impact, while \textit{Formatting Noise} affects specific retrievers and LLMs differently, offering valuable insights for developing RAG-tailored OCR solutions and noise robust models.


\noindent\textbf{Contributions.} We summarize our main contributions:
\begin{itemize}
    \item We present \benchopt, a question-answering benchmark designed to evaluate the impact of OCR on RAG systems. \benchopt includes various unstructured PDF documents from seven RAG domains with ground truth structure data annotations and Q\&A pairs spanning multiple RAG tasks with diverse source of evidences, posing challenges to the employment of current OCR solutions in RAG systems.
    

    
    \item We conduct a comprehensive evaluation of current OCR solutions and reveal that none of them is competent for constructing high-quality knowledge bases for RAG systems. 

    \item We identify two primary types of OCR noise, including \textit{Semantic Noise} and \textit{Formatting Noise}, generate perturbed data with varying levels of noise and explore the trend relationship between the degree of OCR noise and RAG performance.
\end{itemize}

\section{Related Works}
\label{sec:related_works}
\subsection{Retrieval-Augmented Generation}
Retrieval-Augmented Generation (RAG)~\cite{lewis2020retrieval,izacard2023atlas,ram2023context} integrates external knowledge into large language models (LLMs) to mitigate hallucinations. 
Although RAG technology enhances the generation capabilities of LLMs, it is notably sensitive to input noise.
InfoRAG~\cite{xu2024unsupervised} characterizes this noise in RAG as incorrect and irrelevant content within retrieved text and reveals its impact on RAG performance.
RAAT~\cite{fang2024enhancing} further expands noise into relevant noise, counterfactual noise, and irrelevant types.
However, these studies focus solely on chunk-level noise introduced during the retrieval stage and its effect on the generation capabilities of LLMs, leaving the impact of noise derived from OCR results unexplored.
GARAG~\cite{cho2024typos} examines typographical errors, a form of OCR noise, but its scope is limited to plain text using only synthetic data, overlooking the variety of OCR noise encountered in real-world RAG applications.
In this paper, we reveal the impact of noise introduced during the OCR stage, offering a comprehensive analysis of its impact on RAG systems.

\subsection{Document parsing with OCR}
OCR-based document parsing is a promising solution for structured data extraction from unstructured documents, facilitating applications~\cite{zhang2024document,liu2025shifting,luo2024llm} like RAG.
Current OCR solutions can be summarized into three categories, pipeline-based systems~\cite{wang2024mineru,marker,wang2024unimernet}, end-to-end models~\cite{wei2024general,blecher2023nougat,liu2024focus}, and employing VLMs for OCR~\cite{chen2024far,wang2024qwen2,hurst2024gpt,an2024mc,an2025unictokens}
Pipeline-based systems decompose OCR into multiple subtasks, such as layout detection, text, formula, and table recognition, enabling fine-grained data extraction.
End-to-end models take document images as input and output the overall recognition result in an end-to-end manner.
Due to the achievement of VLMs on visual understanding, recent works have explored its application in OCR~\cite{liu2024ocrbenchhiddenmysteryocr,wen2025token}.
In this paper, we evaluate these OCR paradigms, examining their suitability for RAG applications across diverse, real-world document domains.

\subsection{Benchmark and evaluation of Retrieval-Augmented Generation}
Frameworks like RAGAS~\cite{es2023ragas} and ARES~\cite{saad2023ares} propose evaluating RAG systems based on context relevance, answer faithfulness, and answer relevance, using LLMs or fine-tuned discriminators for measurement.
RGB~\cite{chen2024benchmarking} assesses the noise robustness, negative rejection, information integration, and counterfactual robustness of RAG in news data.
MultiHop-RAG~\cite{tang2024multihop} focuses on multi-hop reasoning capabilities, while ClashEval~\cite{wu2024faithful} explores the context preference in conflicting evidence scenarios.
However, these evaluations target specific components of RAG systems, and none of them discusses the impact of external knowledge base construction on RAG systems.
Although UDA~\cite{hui2024uda}, VisRAG~\cite{yu2024visrag} and M3DocRAG~\cite{cho2024m3docrag} explore RAG's effectiveness in document understanding, they consider limited OCR solutions, ignoring challenging documents and lacking analysis of different OCR noise types.
In this paper, we introduce \benchopt to comprehensively investigate OCR noise's impact on RAG systems.


\begin{figure*}[t]
    \centering
    \begin{minipage}[t]{0.23\textwidth} 
        \vspace{0pt} 
        \centering
        \scalebox{0.65}{
            \begin{tabular}{@{}ll@{}}
            \toprule
            \textbf{Statistic} & \textbf{Number} \\
            \midrule
            \textbf{Documents} & $ 1,261 $ \\
            - Domain & $ 7 $ \\
            - Total Pages & $ 8,561 $ \\
            - Avg.Tokens & $ 1,034/\text{page} $ \\
            - Avg.Data Type & $ 1.9/\text{page} $ \\
            \midrule
            \textbf{Questions} & $8,498$\\
            Avg.Question Token & $17.9\pm 8.7$\\
            Avg.Answer Token & $3.5\pm4.1$\\
            \midrule
            \multicolumn{2}{l}{(Evidence Source)}\\ 
            - Text & $3,528\ (41.5\%)$ \\
            - Table & $2,364\ (27.8\%)$ \\
            - Formula & $1,267\ (14.9\%)$ \\
            - Chart & $768\ (9.0\%)$ \\
            - Reading Order & $691\ (8.1\%)$ \\
            \midrule
            (Answer Format) & \\
            - String & $4,171\ (54.4\%)$ \\
            - Numeric & $3,004\ (35.3\%)$ \\
            - Yes/No & $594\ (7.0\%)$ \\
            - List & $483\ (6.9\%)$ \\
            \midrule
            (Task Type) & \\
            - Understanding & $6,114\ (71.9\%)$ \\
            - Reasoning & $2,384\ (28.1\%)$ \\
            - Single-page & $7,656\ (90.1\%)$ \\
            - Multi-page & $842\ (9.9\%)$ \\
            \bottomrule
            \end{tabular}
        }
        \vspace{-2mm}
        \captionof{table}{Dataset Statistics}
        \label{tab:dataset_statistics}
    \end{minipage}%
    \begin{minipage}[t]{0.73\textwidth} 
        \vspace{0pt} 
        \centering
        \includegraphics[width=\linewidth]{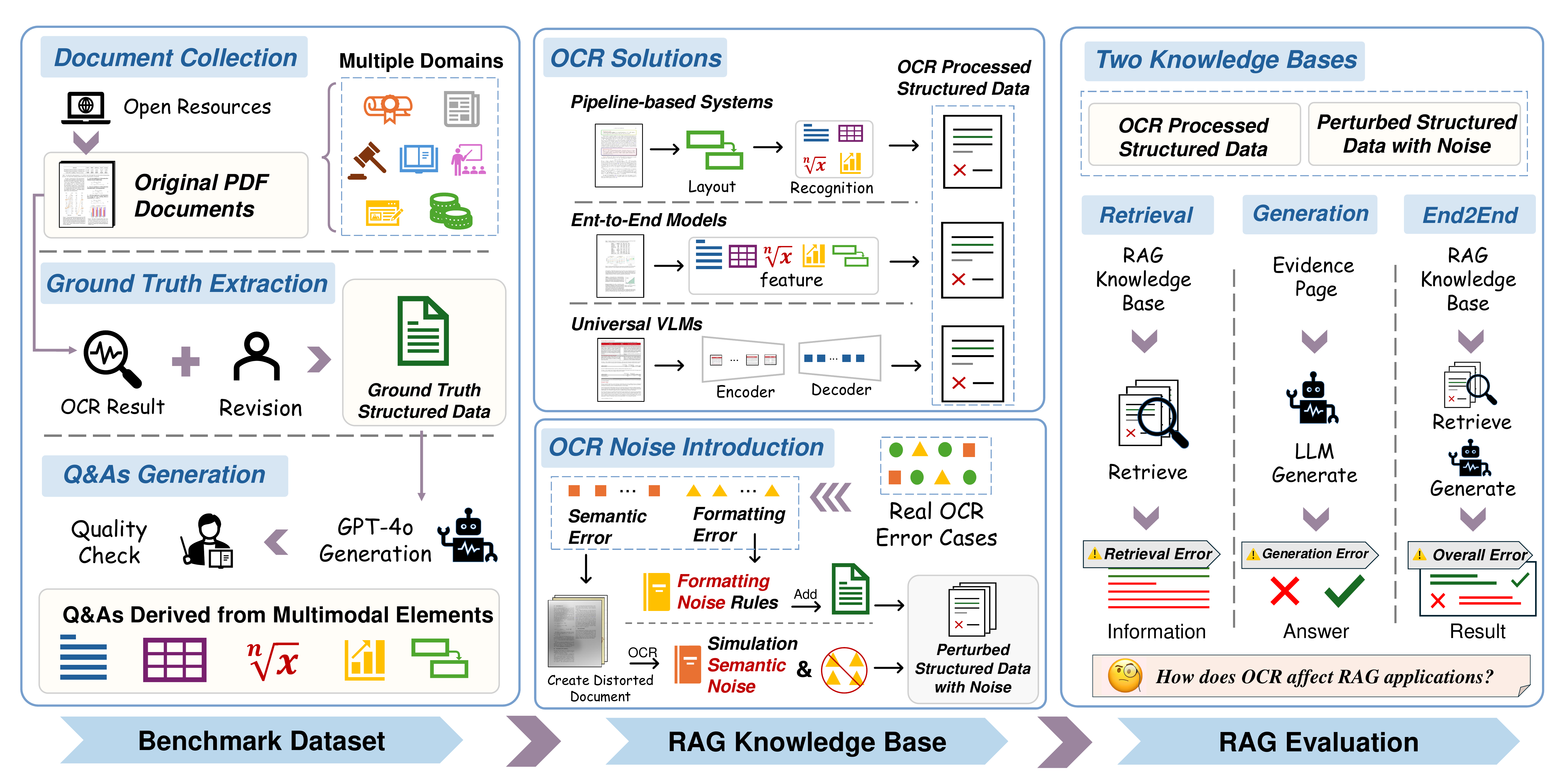}
        \vspace{-6.5mm}
        \caption{Construction of \benchopt and evaluation protocol. (1) Benchmark Dataset: documents from seven domains, human-verified ground truth structured data, and Q\&As from multimodal document elements. (2) RAG Knowledge Base: Current OCR results for benchmarking and perturbed data for assessment. (3) Evaluation of OCR impact on each component and the overall RAG system.}
        \label{fig:pipeline}
    \end{minipage}
    \vspace{-6mm}
\end{figure*}

\section{\benchopt}
Our \benchopt consists of 1) a number of unstructured PDF documents from seven real-world RAG applications, Q\&A pairs derived from multimodal document elements and ground truth structure data annotation, and 2) perturbed structured data based on ground truth with varying degrees of OCR noises.
~\cref{fig:pipeline} illustrates the construction of \benchopt.
We will now delve into the details of each component.

\subsection{Data collection}
\label{method:data_collection}
According to~\cite{zhang2024document,li2024readoc}, extracting structured data from multimodal document elements like formulas and tables poses significant challenges to current OCR solutions. 
Considering the practical application scenarios of RAG and the challenging field of OCR, we compile a PDF document collection representing seven common RAG application scenarios: \textit{Textbook, Law, Finance, Newspaper, Manual, Academic and Administration}.
This collection includes a diverse array of documents from both existing datasets and public web resources.
Specifically, we first collect PDF documents from a wide range of existing datasets, including DUDE~\cite{van2023document}, OmniDocBench~\cite{ouyang2024omnidocbenchbenchmarkingdiversepdf}, FinanceBench~\cite{islam2023financebench}, CUAD~\cite{hendrycks2021cuad}, and GNHK~\cite{Lee2021}.
This results in a highly diverse PDF dataset that encompasses complex structured data and layouts, high text density, handwritten, scanned, and historical documents, as well as multilingual content (Chinese and English), which covers most challenges faced by OCR in document parsing.
In addition to existing datasets, we further supplement our collection with documents from public resources to balance the distribution.
We filter out the corrupted or license-restricted documents and finally curate a document dataset comprising 1,261 PDFs and 8,561 images.
For each collected document, we manually categorize them into 7 domains and provide ground truth structured data.
Specifically, we begin with parsing all documents using Mathpix\footnote{\url{https://mathpix.com/}} for structural data extraction.
We then ask expert-level annotators to revise the results, ensuring fidelity to the original structure and content of PDFs while mitigating any style deviations from Mapthix. 
Detailed descriptions of our selection and processing pipeline can be found in the Appendix ~\cref{appendix:benchmark_details}.

\subsection{Q\&A pairs generation}
\label{method:qa_generation}
The process of extracting structured data from documents involves three key tasks: recognizing plain text; extracting multimodal document elements, including tables, formulas, and charts; and restoring reading order which includes multi-column and truncated paragraph merging.
To systematically assess the impact of OCR results on RAG performance, our Q\&A generation approach revolves around these 5 evidence sources and various realistic Q\&A tasks.
Specifically, we provide the ground truth structured data of each document page to GPT-4o and prompt it to generate Q\&A based on important components in document parsing, including plain text, tables, formulas, and charts.
For questions related to reading order, we identify paragraphs that require merging and instruct GPT-4o to create questions that necessitate combining these paragraphs for a complete answer.
We generate both understanding questions, which only require extracting specific information, and reasoning questions, which involve arithmetic operations, comparisons, or synthesizing information across multiple sections.
For multi-page Q\&A, we derive them from both single-page Q\&A and ground truth structured data from different pages that share the same entity name, recognized with spaCy~\cite{spacy2}.
Detailed process and the prompt template for the Q\&A generation are provided in Appendix~\cref{appendix:prompt}.
Each Q\&A consists of the following fields: one page of the original PDF document, evidence context from ground truth structured data that provides the answer to the question, type of evidence (plain text, table, formula, chart and reading order), and the question and answer which are both derived from this evidence context.
In this way, these Q\&As can serve as a testbed for evaluating OCR results on multimodal document elements.

\noindent\textbf{Quality Control.} The quality of Q\&A pairs generated by a large language model (LLM) can vary significantly.
To address this issue, we apply three data selection criteria to ensure high-quality Q\&As: (1) compatibility with realistic RAG applications, (2) faithfulness to task definitions and (3) correctness.
We incorporate both heuristic methods and prompting LLM for auto data filtering:
\begin{itemize}
    \item \textbf{Compatibility to RAG Applications.} Questions should be context-independent and not answerable by the model's internal knowledge. 
    We collect keywords from existing context-dependent questions, such as ``according to the document'', as heuristic rules.
    Following~\cite{yu2024visrag}, we also employ LLMs to classify context dependence for further filtering.
    Questions answerable without retrieval are excluded by instructing LLMs to answer without access to the evidence context.

    \item \textbf{Faithfulness to Task Definition.} We ensure questions align with their task definition (reasoning or understanding) using LLMs to judge and that evidence sources match the context using heuristic rules.
    In addition, for multi-page question, we provide single evidence context and use LLMs' responses to filter out answerable questions.

    \item \textbf{Correctness.} We verify the accuracy of both evidence context and answers. In specific, we provide LLMs with oracle evidence contexts and sample answers repeatedly to filter Q\&A below a certain correctness threshold.
\end{itemize}
Finally, we manually check against the criteria to ensure the quality.
The LLMs used in our verification include GPT-4o and DeepSeek-V3~\cite{liu2024deepseek}, where we find that DeepSeek-V3 achieve similar performance compared to GPT-4o on Q\&A verification.
A detailed description can be found in the Appendix~\cref{appendix:qa_verification}.
This multi-step quality control ensures the Q\&A dataset meets diverse evidence source requirements and practical RAG applications.
Ultimately, we filter out 8498 high-quality Q\&As from 15317 candidates.

\subsection{Data perturbation with OCR noise}
\label{sec:noise}
Despite advancements in OCR, real-world applications often encounter document types beyond the training corpus of OCR models, leading to low-quality data extraction.
Additionally, the different structured representations of document elements further introduces noise, impacting RAG performance.
In this paper, we focus on two key types of OCR noises: \textbf{\textit{Semantic Noise}} and \textbf{\textit{Formatting Noise}}.
To quantitatively analyze their effects, we start from errors in current OCR results and generate perturbed data with different noise levels.
We will now delve into the details of each type.

\noindent\textbf{\textit{Semantic Noise}} results from OCR prediction errors that impact the semantics of parsed content, deviating retrievers and LLMs from integrating correct information related to user queries.
To systematically capture realistic \textit{Semantic Noise}, we include diverse perturbation to document images and utilize multiple OCR solutions to perform OCR on these document images, capturing a wide range of real-world \textit{Semantic Noise} as much as possible.
We begin with collecting naturally distorted documents and identifying common degradation patterns, such as background artifacts, watermarks, and structural distortions (e.g., dilation and erosion).
Then, we extend from~\cite{blecher2023nougat}, where its method has been shown to be effective for simulating naturally distorted documents, we refine perturbation strategies through an iterative, cross-validated process involving multiple annotators.
One annotator adjusts distortion parameters and applies them to document images, while a another annotator, who is unaware of the applied modifications, distinguish which document appears artificially altered.
This refinement continues until the perturbations become indistinguishable from real samples.
Through this process, we identify 8 effective perturbation types that balance realism and intensity.
Details and examples are provided in Appendix~\cref{appendix:semantic_noise_intro}.
By varying the number and type of perturbations, we generate 3 distinct datasets with controlled \textit{Semantic Noise} levels. 
We then choose MinerU, GOT, and Qwen2.5-VL to curate 9 perturbed data with diverse appearance of \textit{Semantic Noise}, enabling a systematic evaluation of its impact on RAG performance.

\noindent\textbf{\textit{Formatting Noise}} stems from stylistic commands, such as white space characters for beautifying formulas and bold and italic commands for better readability, and inconsistencies in structured data representations across Markdown, LaTeX, and HTML.
Although irrelevant to semantics, this noise complicates information integration for both retrievers and LLMs.
To assess the impact of \textit{Formatting Noise} on RAG, we identify common OCR-induced formatting inconsistencies and develop heuristic rules to introduce controlled perturbations through additions, removals, and format conversions.
A detailed list of perturbation rules is in Appendix~\cref{appendix:formatting_noise_intro}.
By applying these modifications at varying proportions in ground truth structured data, we create three datasets with different degrees of \textit{Formatting Noise}.
Additionally, we evaluate RAG performance under different structured data formats, comparing retrieval and reasoning consistency across Markdown, LaTeX, and HTML representations.

\begin{table*}[t]
    \centering
    \resizebox{\linewidth}{!}{\begin{tabular}{l|c|cccccc|cccccc|cccccc}
        \toprule
        & OCR & \multicolumn{6}{c|}{Retrieval} & \multicolumn{6}{c|}{Generation} & \multicolumn{6}{c}{Overall}\\
        & \makecell{E.D.$\downarrow$} & TXT$\uparrow$ & TAB$\uparrow$ & FOR$\uparrow$ & CHA$\uparrow$ & RO$\uparrow$ & ALL$\uparrow$  & TXT$\uparrow$ & TAB$\uparrow$ & FOR$\uparrow$ & CHA$\uparrow$ & RO$\uparrow$ & ALL$\uparrow$  & TXT$\uparrow$ & TAB$\uparrow$ & FOR$\uparrow$ & CHA$\uparrow$ & RO$\uparrow$ & ALL$\uparrow$  \\
        \midrule
        Ground Truth                             & -    & 81.2 & 69.6 & 74.8 & 70.3 & 9.8 & 70.0     & 49.4 & 46.0 & 34.0 & 47.0 & 28.2 & 43.9    & 45.0 & 34.6 & 28.0 & 32.9 & 18.7 & 36.1 \\
        \midrule
        \multicolumn{20}{l}{\textit{Pipeline-based OCR}} \\
        \midrule
        MinerU~\cite{wang2024mineru}             & 0.24 & 67.7 & 48.5 & 51.1 & 16.5 & \textbf{5.9} & 50.1    & \textbf{45.9} & 39.3 & 28.6 &  9.7 & \textbf{29.5}& \underline{36.7}& \textbf{41.4} & 28.5 & 23.0 &  9.3 & \textbf{17.8}& \underline{30.0} \\
        Marker~\cite{marker}                     & 0.28 & \textbf{75.2} & \underline{57.8} & \underline{55.4} & 19.7 & \textbf{5.9} & 56.6    & \underline{44.5} & 37.8 & 27.8 & 10.9 & \underline{26.2}& 35.9    & 40.1& 28.1 & 22.3 & 10.0 & \underline{16.2}& 29.5 \\
        \midrule
        \multicolumn{20}{l}{\textit{End-to-end OCR}} \\
        \midrule
        GOT~\cite{wei2024general}                & 0.27 & 62.1 & 41.0 & 48.7 &  17.4 & 3.7 & 45.4    & 37.5 & 28.5 & 24.1 &  8.5 &  7.1 & 27.8   & 35.3 & 22.9 & 20.1 &  8.2 & 5.3 & 24.6 \\
        Nougat~\cite{blecher2023nougat}          & 0.34 & 59.1 & 32.7 & 44.2 & 11.3 & 4.4 & 40.9    & 36.7 & 22.9 & 22.9 &  6.4 & 6.9 & 25.5   & 33.5 & 18.4 & 19.4 &  5.8 & 3.6 & 14.5 \\
        \midrule
        \multicolumn{20}{l}{\textit{Vision-Language Model for OCR}} \\
        \midrule
        Qwen2.5-VL-72B~\cite{bai2025qwen2}       & 0.18 & \underline{74.6} & \textbf{59.8} & \textbf{59.7} & \underline{38.2} &  5.3 & \textbf{59.2}    & 44.4 & \textbf{42.1}& \textbf{31.8}& \textbf{27.0}& 11.6 & \textbf{37.5}& \underline{40.6}& \textbf{31.1}& \textbf{26.1}& \underline{19.0}& 8.8 & \textbf{31.1}\\
        InternVL2.5-78B~\cite{chen2024expanding} & 0.28 & 68.2 & 57.7 & 55.3 & \textbf{45.1} &  2.7 & \textbf{55.8}    & 41.8 & \underline{41.8}& \underline{29.0}& \textbf{33.6}&  3.3 & 35.8   & 38.2 & \underline{31.0}& \underline{23.3}& \textbf{22.9}& 3.1 & 29.6 \\
        \bottomrule
        \end{tabular}
    }
    \vspace{-2mm}
    \caption{Evaluation of various OCR solutions and their impacts on RAG systems.
    The OCR performance is reported using edit distance (E.D.).
    We report the generalized LCS or F1 of five types of evidence sources, including plain text (TXT), table (TAB), formula (FOR), chart (CHA), and reading order (RO).
    \textbf{Bold} indicates the best performance, and \underline{underline} indicates the second-best performance.}
    \vspace{-5mm}
    \label{tab:benchmark_ocr}
\end{table*}

\section{Experiments}
\subsection{Experimental settings}
We evaluate the impact of OCR on RAG systems in three ways: retrieval performance, generation performance, and overall system performance.
For the retrieval stage, we utilize knowledge bases derived from the same domains of user queries and retrieve the top-2 matched chunk.
During the generation stage, we provide the page where the question is derived from for LLMs to generate the response.
In the overall evaluation, retrievers retrieve the relevant chunks from the knowledge base in the same domain as the question, and LLMs generate responses based on these chunks.
In the overall evaluation, we provide the top-2 matched chunk for generation unless otherwise stated.
The default chunk size is $1024$ with no overlap.

\noindent\textbf{Metrics.} To evaluate the quality of OCR results, we calculate the edit distance between each page of OCR results and the ground truth structured data and report the average values. 
For assessing retrieval performance, as results of different OCRs often include various extraneous characters, discriminating whether the evidence exactly appears in the retrieved contents is not fair. 
Following~\cite{hui2024uda}, we employ Longest Common Subsequence (LCS)~\cite{paterson1994longest} to measure evidence inclusion in retrieved content.
For the generation stage, we employ the F1-score metric to measure the accuracy of LLMs' responses.

\noindent\textbf{Retrievers.} We consider two primary retrievers: (1) BGE-M3~\cite{bge-m3}, a recent SOTA dense retriever within its size category.
(2) BM-25~\cite{robertson1995okapi,trotman2014improvements} is a lightweight sparse retriever ranking document based on the query term frequency.

\noindent\textbf{LLMs.} We employ three representative open-source LLMs: Qwen2 (Qwen2-7B-Instruct and Qwen2-72B-Instruct)~\cite{wang2024qwen2} and Llama-3.1 (Llama3.1-8B-Instruct)~\cite{dubey2024llama}.
A standard prompt template is used to format responses consistently across all LLMs (see Appendix~\cref{appendix:prompt}).
All open-source models are downloaded from Huggingface~\footnote{\url{https://huggingface.co/}}, with inference conducted on 8 NVIDIA A100 GPUs.

\subsection{Benchmarking current OCR solutions}
\label{sec:exp_bench_ocr}
In this section, we evaluate the suitability of current OCR solutions for real-world RAG applications by conducting comprehensive experiments with our \benchopt.
We involve several representative OCR solutions including (1) Pipeline-based OCR, such as MinerU~\cite{wang2024mineru} and Marker~\cite{marker}, (2) End-to-end OCR, including GOT~\cite{wei2024general} and Nougat~\cite{blecher2023nougat}, and (3) Vision-Language Models, specifically Qwen2.5-VL-72B~\cite{qwen2.5} and InternVL2.5-78B~\cite{chen2024expanding}.
For GOT, we employ its format OCR mode to output structured data.
For Qwen2.5-VL-72B and InternVL2.5-78B, we prompt them to produce formulas, tables, and charts in LaTeX format, with the prompt template available in Appendix~\cref{appendix:prompt}.
The retrievers used are BGE-M3 and BM25, while the LLMs are LLama-3.1-8B-Instruct and Qwen2-7B-Instruct.
All metrics are averaged across domains and combinations of retrievers and LLMs.
Details of experimental results are available in Appendix~\cref{appendix:additional_exp}.

Through the comparison presented in~\cref{tab:benchmark_ocr}, we derive several key conclusions about the performance of these OCR solutions and their corresponding impacts on RAG systems, as follows:
\textbf{(1) VLMs for OCR achieve the best overall performance.}
Among all OCR solutions, Qwen2.5-VL-72B consistently outperforms others across all three evaluation stages.
Its superiority stems from its ability to handle structured data more effectively than both pipeline-based and end-to-end OCR methods.
Despite claims that GOT can parse charts, its performance remains subpar.
Similarly, MinerU, Marker, and Nougat fail to produce comparable results due to their inability in parsing chart.
For plain text questions, although its poor performance on high-resolution documents (see newspaper in APPendix~\cref{appendix:additional_exp}, Qwen2.5-VL-72B performs comparably to pipeline-based OCR.
Our manual review suggests that its strong language decoder enhances robustness against historical and distorted documents, a capability lacking in pipeline-based OCR.
\textbf{(2) Reading order is challenging for VLMs and End-to-end OCR.}
Despite their strong semantic understanding capabilities offered by language decoders, both VLM and end-to-end OCR struggle with merging paragraphs correctly, reflected in F1-score of just 8.8 in overall evaluation.
In contrast, pipeline-based OCR, though lacking semantic understanding, achieves performance close to ground truth (GT) using rule-based strategies.
However, GT itself performs poorly, likely because reading-order questions require integrating information across multiple paragraphs, posing challenges for current RAG systems~\cite{tang2024multihop}.
\textbf{(3) All OCR solutions exhibit performance degradation.}
Even the best solutions experience a 14\% (5 F1-score) drop in the overall stage evaluation, with greater losses in the retrieval and generation stages. 
This indicates that our \benchopt presents significant challenges for both OCR solutions and RAG systems.

In summary, current OCR solutions struggle to maintain robustness and effectiveness across diverse real-world RAG application scenarios. 
Additionally, standard OCR metrics like edit distance do not always align with RAG performance.
For example, while MinerU and Qwen2.5-VL-72B exhibit lower edit distances compared to Marker, they do not consistently achieve better performance across all metrics. 
This discrepancy may be attributed to the varying types of OCR noise introduced by different solutions. 
To further investigate, we systematically explore the impact of these OCR noise types on RAG in~\cref{sec:exp_ocr_noise}.

\subsection{In-depth analysis of OCR noise's impact on RAG}
\label{sec:exp_ocr_noise}

\begin{figure*}[t]
    \centering
    \vspace{-3mm}
    \includegraphics[width=0.92\linewidth]{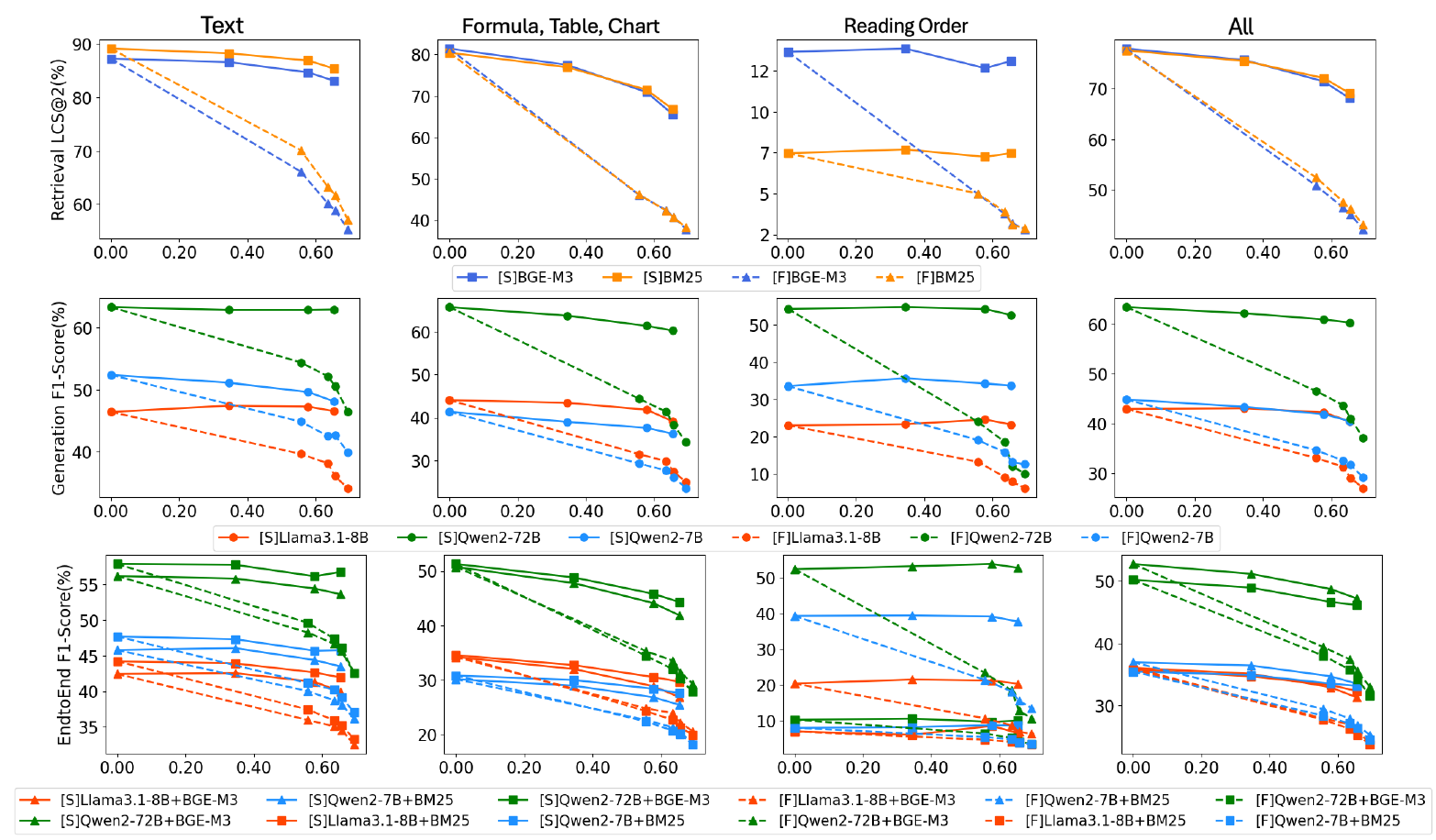}
    \vspace{-2.5mm}
    \caption{Impact of \textit{Semantic Noise} ([S] dashed lines) and \textit{Formatting Noise} ([F] solid lines) on RAG components.
    The horizontal axis denotes the ratio $r_{\text{noise}}$, where higher values indicate greater OCR-induced noise.
    We report LCS and F1-score for each evidence source: text (first column), the average score for multimodal elements (tables, formulas, and charts, second column), reading order (third column), and all sources combined (last column).
    }
    \vspace{-7mm}
    \label{fig:exp_noise}
\end{figure*}

In this section, we conduct an in-depth analysis of the impact of \textit{Semantic Noise} and \textit{Formatting Noise} on RAG systems, using perturbed structured data with varying levels of perturbations.
For each type of OCR noise, we introduce three noise levels—mild, moderate, and severe—to systematically assess their effects.
Since our findings in~\cref{sec:exp_bench_ocr} indicate that edit distance fails to accurately capture the degree of OCR noise, we instead define the ratio $r_{\text{noise}}$, representing the proportion of Q\&A pairs affected by OCR noise as a measure.
Specifically, for each Q\&A pair, we compute the LCS between its evidence context and the corresponding perturbed structured data.
If LCS exceeds 0.95, the Q\&A pair is considered as unaffected, otherwise, it is affected.
We then use the ratio ($r_{\text{noise}}$) to quantify the degree of perturbations, with 0 representing to ground truth structured data and values approaching 1 indicating greater perturbation.
This approach allows us to align the perturbation levels of \textit{Formatting Noise} and \textit{Semantic noise} for fair comparisons.
Additionally, for \textit{Formatting Noise}, we evaluate its retrieval performance with modified LCS calculations by excluding stylistic commanding introduced during perturbation, ensuring a fair assessment of retrieval accuracy.
For each degree of \textit{Semantic Noise}, we report the average RAG performance using three different OCR results, including MinerU, GOT, and Qwen2.5-VL-72B.

\subsubsection{Fine-grained impact on retrieval and generation}
\noindent\textbf{\textit{Semantic Noise} significantly influences both retrieval and generation phases.}
As illustrated in~\cref{fig:exp_noise}, increasing \textit{Semantic Noise} from mild ($r_{\text{noise}}=0$) to severe ($r_{\text{noise}}>0.6$) results in nearly a $50\%$ performance decline for most retrievers and LLMs.
In the retrieval stage, both the sparse retriever BM25 and the dense retriever BGE-M3 suffer consistent performance declines across all types of questions, suggesting that dense retrieval's stronger comprehension does not provide robustness against \textit{Semantic Noise}.
In the generation phase, all LLMs struggle with \textit{Semantic Noise}, among which performance on reading-order decreases the most.
Interestingly, although the way we introduce \textit{Semantic Noise} should primarily affect text recognition, questions related to multimodal elements (tables, formulas and charts) degrade even further, highlighting the challenges in parsing, understanding and reasoning over multimodal document data.

\noindent\textbf{\textit{Formatting Noise} primarily affects multimodal questions.}
While performance on plain text queries and reading-order-related questions remain largely unaffected, retrieval and generation performance drops more severe for multimodal queries.
The maximum performance losses reach 12.7\% for BGE-M3 and 9.1\% for Llama3.1-8B in retrieval and generation, respectively.
In addition, larger LLMs exhibit greater robustness, with only a 7\% performance reduction on multimodal questions, indicating that more advanced LLMs can effectively handle \textit{Formatting Noise}.

\subsubsection{Impact on end-to-end evaluation}
\noindent\textbf{\textit{Semantic Noise} consistently demonstrate a strong impact, while \textit{Formatting Noise} affects specific retrievers and LLMs differently.}
\textit{Semantic Noise} consistently degrades performance across all combination of retrievers and LLMs, particularly on multimodal questions involving tables, formulas, and charts.
In contrast, the effect of \textit{Formatting Noise} is more variable when using smaller LLMs, Llama3.1-8B and Qwen2-7B, for generation, despite greatly reduced retrieval accuracy, the overall performance shows a slight change due to their limited information integration capabilities.
Conversely, using a larger LLM, Qwen2-72B, is highly sensitive to retrieval performance. 
While its generation performance shows light changes, the overall performance decreases, especially on questions related to multimodal elements.

In summary, \textit{Semantic Noise} significantly affects each stage of RAG and the entire system.
The impact of \textit{Formatting Noise} varies with different retrievers and generators, particularly affecting questions with multimodal elements.

\begin{figure}[t]
    \centering
    \includegraphics[width=\linewidth]{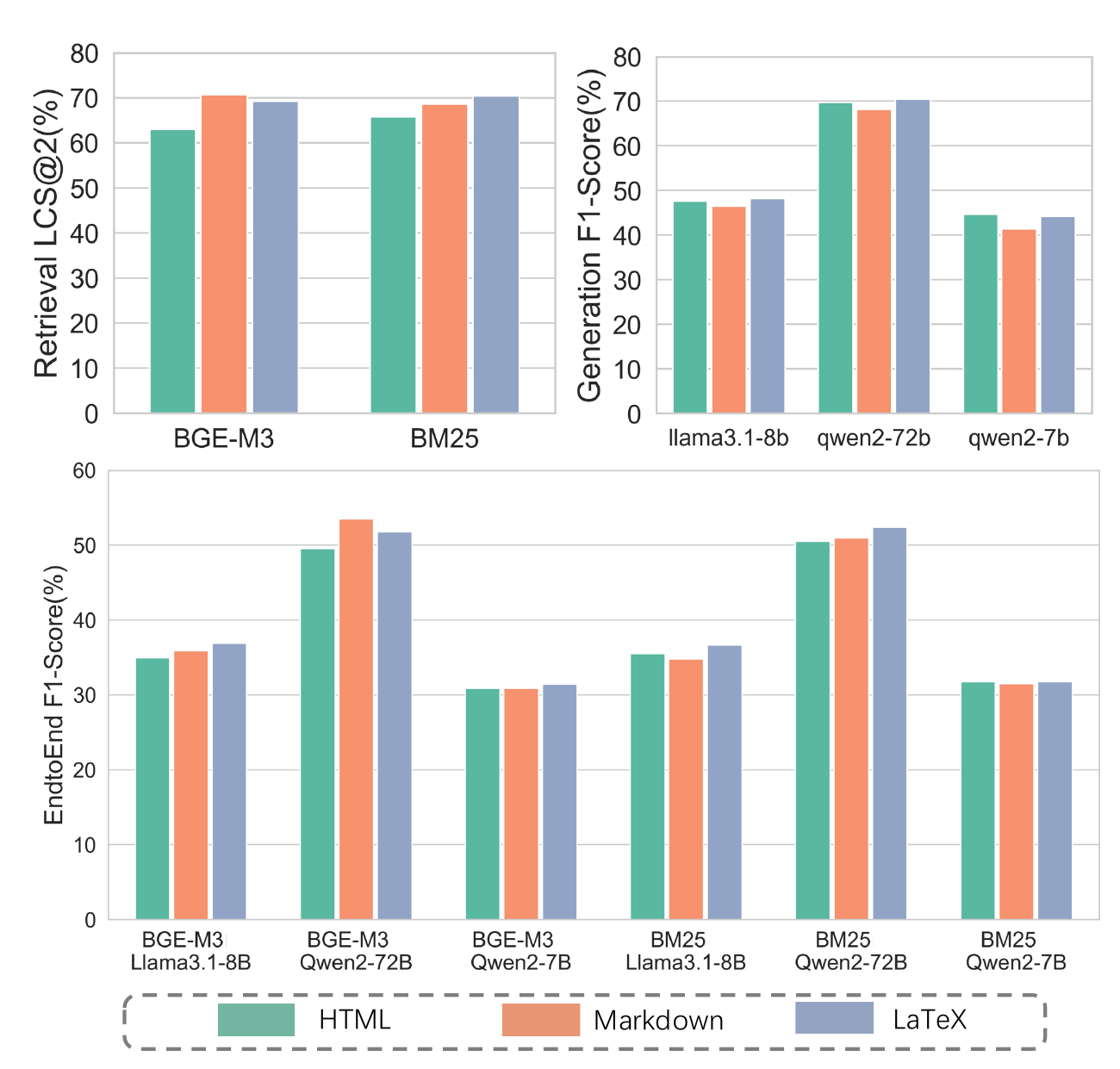}
    \vspace{-5.5mm}
    \caption{Performance of retrieval, generation and end-to-end with different table format. We only report the results of table-related questions.}
    \vspace{-6mm}
    \label{fig:tab_format}
\end{figure}

\subsubsection{Impact of table format}
In addition to perturbations, we investigate the influence of different table formats as a kind of \textit{Formatting Noise}.
As shown in~\cref{fig:tab_format}, HTML tables show inferior performance during retrieval compared to the Markdown and LaTeX formats.
Markdown and LaTeX formats perform similarly, with BGE-M3 demonstrating a better understanding of Markdown.
In the generation phase, HTML and LaTeX showed similar performance across all models, but the Markdown format performed worse due to the lack of support for merging cells.
In end-to-end evaluations affected by low retrieval performance, using HTML tables is comparatively worse, while the combination of Qwen2-72B and BGE-M3 using Markdown achieves the best performance.

\subsubsection{Error analysis}
\begin{figure}
    \centering
    \includegraphics[width=\linewidth]{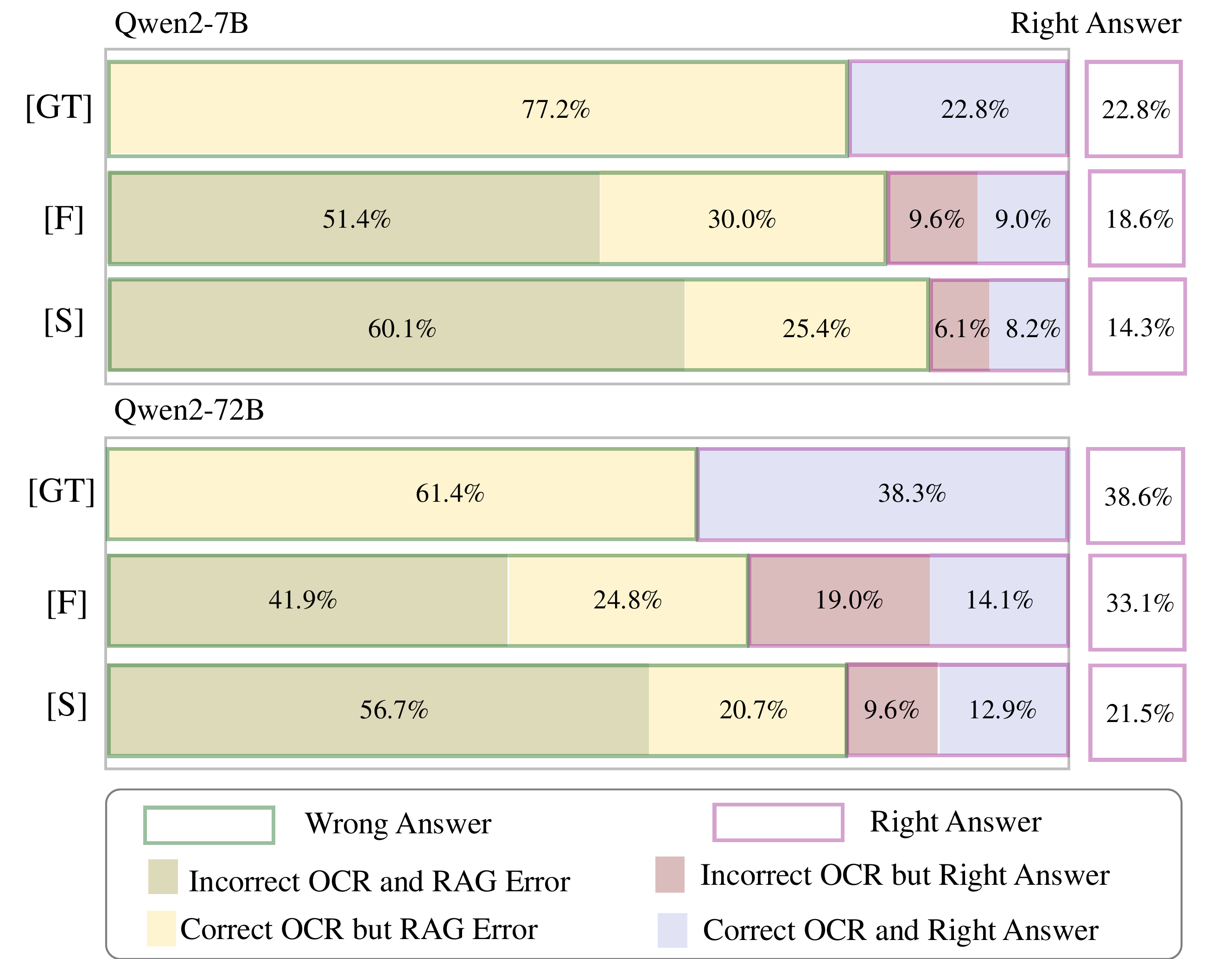}
    \vspace{-4mm}
    \caption{Analysis of answer correctness distribution in Q\&A pairs, using different perturbed data and LLMs. 
    It reveals that larger LLMs are more robust to OCR noise.
    [S] and [F] denote perturbed data with severe \textit{Semantic Noise} and \textit{Formatting Noise}, respectively.
    [GT] represents ground truth perturbed data.
    }
    \vspace{-4.5mm}
    \label{fig:error_analysis}
\end{figure}

We further conduct error analysis to understand the bottlenecks of the RAG system with OCR noise in a quality approach.
Specifically, we calculate the distribution of OCR errors and RAG errors when using ground truth structured data and perturbed data with severe \textit{Formatting Noise} and \textit{Semantic Noise}.
Our evaluation employs BGE-M3 as the retriever and assesses error distributions using two generators: Qwen2-7B and Qwen2-72B.
We use the same strategies to identify the proportion of OCR errors as in~\cref{sec:exp_ocr_noise}.
The distribution of these errors is illustrated in~\cref{fig:error_analysis}.
It indicates that when the proportion of OCR errors is similar for \textit{Semantic Noise} and \textit{Formatting Noise} (66\% vs. 61\%), it performs better with \textit{Formatting Noise}.
Of these, about half of the correct responses when using Qwen2-7B as a generator were done by the model with incorrect OCR results.
Meanwhile, larger LLMs are more robust to OCR noise.
Compared to Qwen2-7B, Qwen-72B has a 9.4\% and 3.5\% higher percentage of samples with OCR errors but ultimately correct in \textit{Formatting Noise} and \textit{Semantic Noise}, respectively.

\section{Conclusion}
In this paper, we present \benchopt to evaluate the impact of OCR on RAG systems, which encompasses diverse PDF documents from sven RAG application scenarios along with Q\&A pairs derived from multimodal elements in these documents.
Through comprehensive evaluations of current OCR solutions, we reveal that none is fully capable of RAG systems across all scenarios.
Furthermore, our analysis of different types of OCR noise demonstrates that while no retrievers and LLMs are immune to \textit{Semantic Noise}, more advanced models exhibit greater resilience to \textit{Formatting Noise}.
We believe that our documents featuring challenging OCR attributes and Q\&A pairs sourced from varied document elements, will advance the development of OCR solutions tailored for RAG and OCR noise-resistant RAG systems.

\section*{Acknowledgments}
This work is supported by Shanghai Artificial Intelligence Laboratory, the National Key R\&D Program of China (2024YFA1014003), the National Natural Science Foundation of China (Grants 92470121 and 62402016), and the CAAI-Ant Group Research Fund.

\medskip

{
    \small
    \bibliographystyle{ieeenat_fullname}
    \bibliography{main}
}

\clearpage
\setcounter{page}{1}
\maketitlesupplementary

\setcounter{section}{0}
\setcounter{table}{0}
\setcounter{figure}{0}
\renewcommand{\thesection}{\Roman{section}}
\renewcommand{\thetable}{S\arabic{table}}
\renewcommand{\thefigure}{S\arabic{figure}}

\newcommand{\tb}[1]{\textbf{#1}}
\newcommand{\ud}[1]{\underline{#1}}

\section{Instruction Prompts}
\label{appendix:prompt}

\noindent\textbf{Q\&A Generation Prompt Template.}
The template is shown in~\cref{tab:qa_gen_prompt}. 
Following~\cite{wu2024faithful,zou2024docbench}, we instruct GPT-4o to generate questions with clear entities and require three levels of difficulty for question diversity.

\noindent\textbf{RAG Generation Stage Prompt Template.}
The prompt template for LLMs and VLMs with text-only input is shown in~\cref{tab:llm_prompt}. 

\noindent\textbf{Vision-Language Models OCR Prompt Template.}
We tune the prompt for the best performance of VLMs OCR, by comparing simple and detailed instructions as shown in~\cref{tab:ocr_prompt}.
Results in~\cref{tab:ocr_prompt_comp} indicate that the detailed prompt consistently performs better across all evaluations, so it is used by default.

\section{Benchmark Construction Details}
\label{appendix:benchmark_details}
\subsection{Document details}
We curate a dataset of 1,261 PDFs spanning 8,561 pages, with 3,596 pages designated for Q\&A generation and the remainder forming part of the knowledge base.
These PDFs are sourced from DUDE~\cite{van2023document}, OmniDocBench~\cite{ouyang2024omnidocbenchbenchmarkingdiversepdf}, FinanceBench~\cite{islam2023financebench}, CUAD~\cite{hendrycks2021cuad}, GNHK~\cite{Lee2021}, and public resources, including Arxiv\footnote{\url{https://arxiv.org}}, ManualsLib\footnote{\url{https://www.manualslib.com/}}, LibreTexts\footnote{\url{https://libretexts.org/}}.

\noindent\textbf{DUDE:} We extract documents from the validation and test splits of DUDE, applying manual screening based on our criteria~\cref{fig:overview} to exclude samples infeasible for structured data parsing and classify each of them into 7 domains.
We finally selected 450 PDFs with 4,058 images from 2,069 PDF candidates.

\noindent\textbf{OmnidocBench:} OmniDocBench~\cite{ouyang2024omnidocbenchbenchmarkingdiversepdf} features span-level annotations and presents challenges for OCR due to its multilingual, high-resolution with dense text and handwritten content.
We select all newspaper documents and manually review textbook-related samples, eliminating those with low knowledge density that hinder meaningful Q\&A generation. This process yields 289 PDFs.

\noindent\textbf{FinanceBench:} Following prior observations~\cite{ma2024mmlongbench}, dboth DUDE~\cite{van2023document} and FinanceBench~\cite{islam2023financebench} contain diverse document types.
From FinanceBench, we randomly sample 10 PDFs characterized by large, complex tables and charts.

\noindent\textbf{CUAD:} We randomly select 65 PDFs to supplement the documents in law domains, which all have high text density.

\noindent\textbf{GNHK:} GNHK consists of handwritten documents. We manually assess and remove those with low knowledge density, finalizing a selection of 172 PDFs.

Each document is manually reviewed by primary authors to ensure its availability for academic use.
Detailed domain statistics are shown in~\cref{tab:supp_doc_page_statistic}
\begin{table}[h!]
    \centering
    \begin{tabular}{lccc}
    \toprule
    Domains & PDFs & Pages & Pages with Q\&As \\
    \midrule
    Law             & 95 & 1187 & 1143 \\
    Finance         & 65 & 2133 & 1359 \\
    Textbook        & 504& 678  & 1126 \\
    Manual          & 87 & 1724 & 1155 \\
    Newspaper       & 279& 487  & 1202 \\
    Academic        & 85 & 1011 & 1181 \\
    Administration  & 146& 1341 & 1332 \\
    Total           & 1261& 8561& 8498 \\
    \bottomrule
    \end{tabular}
    \caption{Document statistics of each domain}
    \label{tab:supp_doc_page_statistic}
\end{table}

\subsection{Ground truth structured data annotation}
We annotate the ground truth structured data using Mathpix Markdown format, where tables and formulas are represented in LaTeX.
Chart data is extracted in LaTeX table format, with charts lacking clear numeric values in figure filtered out.
For images in documents, any parsable text is retained as plain text in the corresponding section.
To ensure high-quality annotations, we first use Mathpix to pre-annotate all PDFs. 
Finally, the primary authors employ Mathpix Markdown previews\footnote{
\url{https://github.com/Mathpix/vscode-mathpix-markdown}} to render structured data into PDFs, manually review and correct pre-annotated results.

\subsection{Document with challenging attributes}
\label{appendix:complex_layout}
Although existing RAG document benchmarks have gathered PDFs from different domains~\cite{hui2024uda,faysse2024colpali,cho2024m3docrag}, they often ignore the challenges posed by OCR. 
To address this, we construct a benchmark that explicitly incorporates documents with challenging attributes.
We define nine key attributes: structured data (tables, formulas, charts), complex layouts, handwritten content, distortions, scanned PDFs, dense text, and multilingual content. 
Structured data, dense text (exceeding 770 tokens), and multilingual pages are classified based on the annotated ground truth structured data. 
A document is considered to have a complex layout if its layout detection yields more than 20 bounding boxes.
Distorted, scanned, and handwritten documents are identified during manual checks.

\subsection{Q\&A generation}
To generate high-quality Q\&A pairs covering diverse tasks and evidence sources, we define multiple prompts for each task, as detailed~\cref{tab:qa_gen_prompt,tab:task_prompt,tab:mp_prompt}.
For Chinese questions, we provide the same set of templates in Chinese to ensure that the model generates Chinese responses.
\noindent\textbf{Q\&A with different evidence sources.} For Q\&A generation with evidence sourced from plaint text, table, formula and chart, we extract relevant pages from the ground truth structured data and use GPT-4o to generate Q\&A pairs grounded in the corresponding evidence via tailored prompts.
For Q\&A related to reading order, we leverage MinerU~\cite{wang2024mineru}, the leading model for reading order recognition~\cite{ouyang2024omnidocbenchbenchmarkingdiversepdf}, to identify the reading order and bounding box of paragraphs in each document.
When working with documents from OmniDocBench~\cite{ouyang2024omnidocbenchbenchmarkingdiversepdf}, we directly use the ground truth reading order from its annotations. 
We verify the layout detection and reading order predictions, selecting paragraph pairs that meet one of the following criteria:
\begin{itemize}
    \item Adjacent paragraphs in reading order whose bounding boxes are not vertically aligned.
    \item Paragraphs separated by multimodal document elements (e.g., block formulas, tables, or images).
\end{itemize}
We then randomly sample 1,500 candidate matches, manually correcting approximately 20\% where MinerU's predictions are inaccurate.
We then prompt GPT-4o to generate Q\&A pairs using the prompts in~\cref{tab:task_prompt}.
We find that this simple prompting-based strategy can effectively generate questions with diverse evidence sources, with over 90\% correctly aligned with their evidence source in our Q\&A verification process.

\noindent\textbf{Q\&A with different tasks.} To generate both understanding and reasoning questions, we apply the corresponding prompts from~\cref{tab:task_prompt}.
For multi-page Q\&A generation, we employ two different approaches to generate Q\&A candidates: 
(1) Combine questions from two single-page Q\&As that mention the same entity. (2) Generating multi-page questions from two paragraphs on different pages that reference the same entity.
Specifically, we use spaCy~\cite{spacy2} for named entity recognition in both single-page Q\&As and document paragraphs.
We then filter out candidate pairs, including: (1) Single-page Q\&A pairs where the entity in one answer appears in another question. (2) Paragraph pairs that share the same entity.
We finally utilize the prompts in~\cref{tab:mp_prompt} to generate multi-page questions.
However, despite the many optimizations of the prompt and generation strategies we tried, GPT-4o sometimes produces Q\&A pairs that are either answerable with a single paragraph or simply concatenate two single-page questions while maintaining separate evidence sources instead of high-quality and realistic multi-page Q\&As.
To address these limitations, we develop a comprehensive filtering process to ensure the quality of multi-page Q\&As, as detailed in~\cref{appendix:qa_verification}.

\subsection{Q\&A verification.}
\label{appendix:qa_verification}
We verify Q\&A quality based on three criteria: (1) Compatibility with realistic RAG applications, (2) faithfulness to task definition, and (3) correctness.
Below, we detail our approach for each aspect.

\noindent\textbf{Compatibility with Realistic RAG Applications.}
To assess context dependence, we identify key patterns from existing context-dependent questions and apply the following heuristics:
\begin{itemize}
    \item Questions lacking an explicit entity name.
    \item Questions containing more than one ambiguous pronouns (e.g., "he," "she," "it," "they", "this", "that").
    \item Questions featuring phrases such as "in the document" or "according to the document."
\end{itemize}
These rules filter most context-dependent questions.
We then refine the selection using prompts in VisRAG~\cite{yu2024visrag} and DeepSeek-V3 to further distinguish context-dependent questions from the remaining set.
Additionally, we use GPT-4o to exclude questions answerable without retrieval by instructing it to respond without providing evidence context across both single-page and multi-page Q\&As.

\noindent\textbf{Faithfulness to Task Definition.} 
Based on the Q\&A verification prompts in~\cite{deng2024longdocurl}, we use the prompts in~\cref{tab:qa_verification_prompt} to assess faithfulness using DeepSeek-V3.
To verify the validity of evidence sources, we locate them in the original ground truth structured data and ensure they originate from the correct corresponding LaTeX code environments.
For the multi-page and reading-order questions, we employ GPT-4o to generate three responses: (1) without context, (2) with context A, and (3) with context B.
If any response yields a correct answer, the question is excluded, ensuring that only truly multi-page or reading-order-related questions remain.

\noindent\textbf{Correctness.}
To guarantee each Q\&A has a unique and correct answer supported by its evidence context, we provide oracle evidence and sample GPT-4o’s response 10 times.
We apply a best-of-N strategy to determine the final answer, which must match the ground truth.
Q\&As with fewer than three consistent correct responses are also excluded.

Our filtering pipeline underwent two iterations of refinement.
In each round, we randomly sample 100 Q\&As to verify the filtering results adherence to our criteria.
Finally, to mitigate false positives, we manually reviewed all remaining questions, yielding 8,498 high-quality Q\&As from an initial pool of 15,317 candidates.

\section{OCR Noise Introduction}
\subsection{Rules for Formatting Noise introduction}
\label{appendix:formatting_noise_intro}
To introduce \textit{Formatting Noise}, we define a perturbation rate $r$ to control its extent.
In order to match the level of \textit{Semantic Noise} (measured by similar edit distance), we set the $r=\{0.1, 0.3, 0.6\}$, indicating the three levels of perturbation: mild, moderate, and severe.
Based on the \textit{Formatting Noise} in the existing OCR results, we formulate the following rules to perturb plain text, tables, and formulas, respectively.

\subsubsection{Plain text}
\noindent\textbf{Text Style}: Given the plain text content of the ground truth, we randomly divide it into a sequence where each item consists of 2 to 5 words, select target items based on $r$, and apply one of the following operations as perturbations.
\begin{itemize}
    \item Bold: Enclose the selected text in \texttt{**} or \texttt{\textbackslash textbf\{\}}.
    \item Italic: Enclose the selected text in \texttt{*}  or \texttt{\textbackslash textit\{\}}.
    \item Underline: Enclose the selected text in \texttt{\_}  or \texttt{\textbackslash underline\{\}}.
\end{itemize}

\noindent\textbf{Title Formatting}: We identify short sentences that end with a full stop and have no more than 5 words as potential headings.
We randomly pick them according to $r$ and add one of level 1 to level 3 title controls in Markdown (\texttt{\#}) or LaTeX (\texttt{\textbackslash section\{\}}) to make new titles.

\noindent\textbf{Paragraph}: To simulate the line breaks that exist in PDFs, we randomly insert \texttt{\textbackslash n} at word intervals based on $r$.

\subsubsection{Formula}
\noindent\textbf{Formula Conversion}: Randomly convert the inline formula into block formula and vice versa at rate $r$.

\noindent\textbf{Extraneous Elements}: We first randomly select the target formulas based on $r$.
Subsequently, for each target formula, we randomly insert 1 to 5 meaningless markers in its symbol gaps, including \texttt{\textbackslash}, \texttt{\textbackslash quad}, \texttt{\textbackslash qquad}, \texttt{\textbackslash;}, \texttt{\textbackslash:}.

\noindent\textbf{Equivalent Symbols}:
For each formula, we replace the following equivalent symbols with probability $r$:
\begin{itemize}
    \item bold: \texttt{\textbackslash mathbf\{\}}, \texttt{\textbackslash boldsymbol\{\}}.
    \item cursive: \texttt{\textbackslash mathbb\{\}}, \texttt{\textbackslash pmb\{\}}, \texttt{\textbackslash mathrsfs\{\}}, \texttt{\textbackslash euscript\{\}}, \texttt{\textbackslash mathcal\{\}}.
    \item unicode: \texttt{(\textbackslash sigma,\textbackslash u03A3)}, etc\footnote{Full lists are drawn from \url{https://raw.githubusercontent.com/w3c/xml-entities/refs/heads/gh-pages/unicode.xml}}.
\end{itemize}

\subsubsection{Table}
\noindent\textbf{Row and Column Lines}: For each line and column, randomly insert \texttt{\textbackslash hline} or \texttt{\textbackslash cline} with probability $r$.

\noindent\textbf{Cell Content}: For each cell content, randomly apply above rules for plain text or formula with probability $r$.

\begin{table}[h!]
    \centering
    \begin{tabular}{lc}
        \toprule
        OCR & Avg. Counts \\
        \midrule
        MinerU            & 35.0 \\
        GOT               & 45.7 \\
        Nougat            & 63.2 \\
        F-minor           & 37.9 \\
        F-moderate        & 42.2 \\
        F-severe          & 56.3 \\
        \bottomrule
    \end{tabular}
    \caption{Counts of \textit{Formatting Noise}. The counts of \textit{Formatting Noise} we add (F-minor, F-moderate, F-severe) is approximately the distribution of the counts of \textit{Formatting Noise} for MinerU, GOT and Nougat.}
    \vspace{-5mm}
    \label{tab:my_label}
\end{table}

\subsection{Rules for Semantic Noise introduction}
\label{appendix:semantic_noise_intro}
\begin{figure}[t]
    \centering
    \includegraphics[width=1\linewidth]{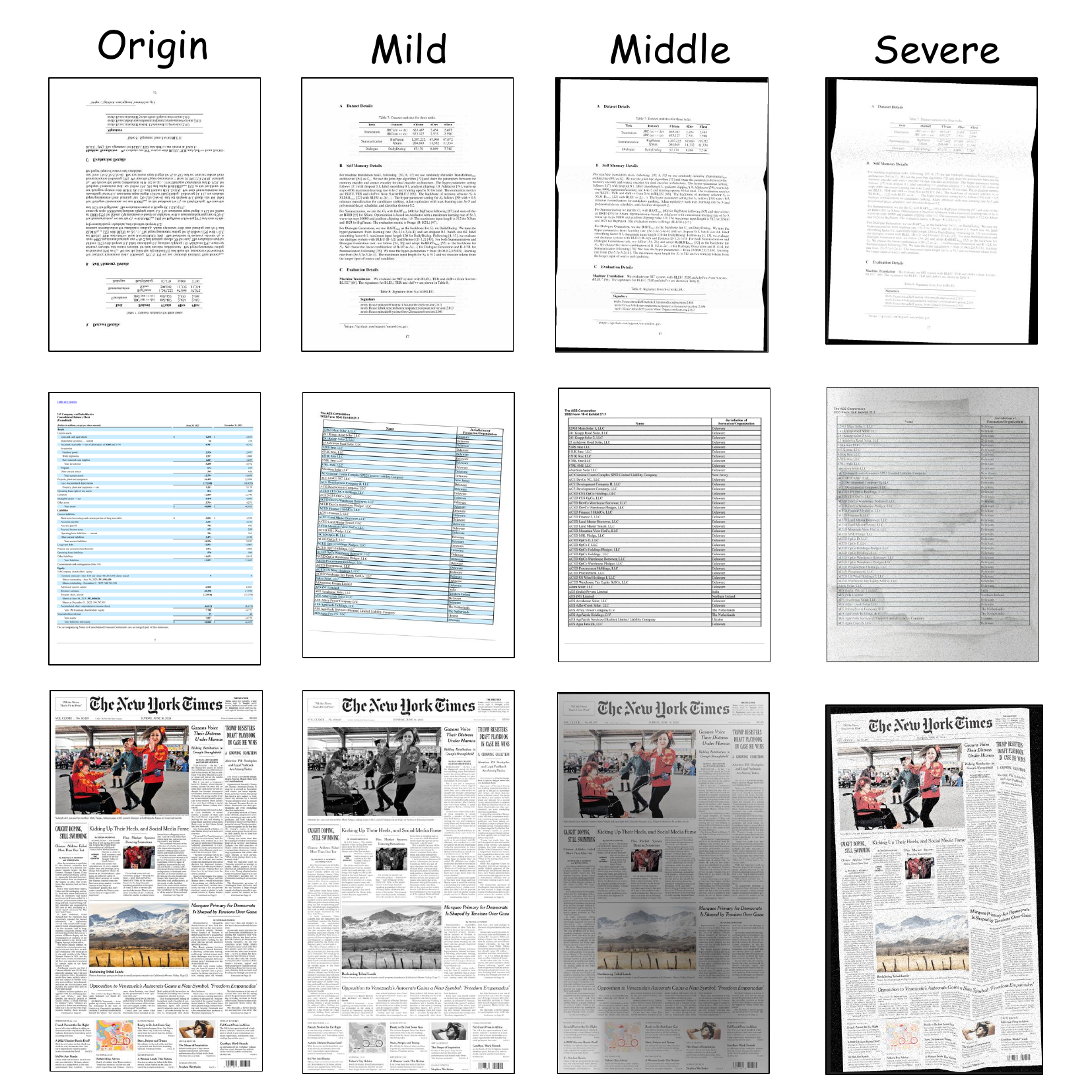}
    \vspace{-5mm}
    \caption{Cases of distorted documents.}
    \label{fig:distorted_documents}
\end{figure}

\begin{figure}[t]
    \centering
    \includegraphics[width=\linewidth]{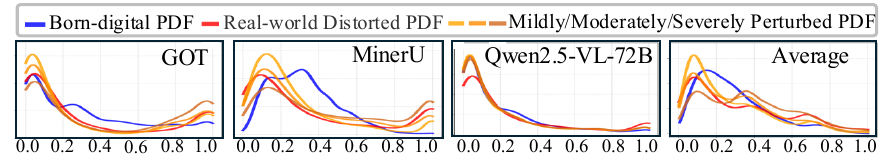}
    \caption{Distribution of \textit{Semantic Noise}.
    X-axis denotes edit distance. Mild/moderate/severe perturbation is based on born-digital PDFs.}
    \label{fig:s_noise_distri}
\end{figure}

In order to construct perturbed document images that conform to the realistic distribution of naturally distorted documents, we use a cross-validated process involving multiple annotators.
We finally identify 8 strategies from~\cite{blecher2023nougat} as follows:
\begin{itemize}
    \item \textbf{Background Addition:} We collect 15 background images of real paper textures and blend them with original images at an 80:20 ratio.
    \item \textbf{Salt-and-Pepper Noise:} Randomly replace 1\% of the image pixels with white ("salt" noise) and black ("pepper") pixels.
    \item \textbf{Dirty Rollers:} Add random rollers with thickness between 1 and 3 pixels, a line addition probability of 0.05 per pixel row or column.
    \item \textbf{Random Rotation:} Apply a random rotation of $-3^{\circ}$ and $+3^{\circ}$.
    \item \textbf{Binarization:} Utilize the Augraphy\footnote{\url{https://github.com/sparkfish/augraphy}} to simulate effects such as dilation, erosion, and letterpress printing.
    \item \textbf{Warping:} Apply geometric transformations and folding effects via Augraphy to mimic paper creases.
    \item \textbf{Shadows:} Apply light gradients and shadow cast from Augraphy to simulate shadows that occur when a document is taken.
    \item \textbf{Blur via Point Spread Function:} Generated 100 PSF kernels and randomly applied one to the document.
\end{itemize}
We classify above distortions into two categories: (1) weak distortions: These preserve text clarity and include background addition, binarization, minor rotation, and PSF-based blurring. (2) strong distortions: These degrade readability, causing blurriness and font warping. They include salt-and-pepper noise, dirty rollers, warping, and shadow effects.
To simulate varying levels of document distortion, we apply the above strategies in three ways:
\begin{itemize}
    \item Apply a weak distortion per page.
    \item Apply a strong distortion per page.
    \item Apply two randomly selected distortions per page.
\end{itemize}
We generate three document image datasets with varying noise levels and parse structured data using MinerU, GOT, and Qwen2.5-VL, resulting in 9 perturbed datasets.
The examples of distorted documents are shown in~\cref{fig:distorted_documents}.
The distribution of introduced \textit{Semantic Noise} is illustrated in~\cref{fig:s_noise_distri}.
In most cases, the distributions of our perturbed PDFs align with those of real-world distorted PDFs, validating the realism of our method.
In~\cref{sec:exp_ocr_noise}, we evaluate RAG performance on these datasets, reporting the average results for each noise level.

\noindent\textbf{Ratio of OCR noise in real-world OCR results.} To illustrate the frequency of OCR noise in real-world OCR results, We match corresponding TXT/FOR/TAB blocks, which includes \textasciitilde 130 tokens each, and show the ratio of \textit{Semantic Noise} and \textit{Formatting Noise} in~\cref{tab:noise_frequency}.
\begin{table}[t]
    \centering
    \resizebox{\linewidth}{!}{
        \begin{tabular}{lcccc}
        \toprule
        \multirow{2}{*}{OCR}         & \multicolumn{3}{c}{SN} & \multirow{2}{*}{FN} \\
                    & TXT & FOR & TAB & \\    
        \midrule
        MinerU      & 40.3\% & 78\%/34\% & 79\%/58\% & 10.9 \\
        Qwen2.5-VL  & 31.6\% & 46\%/25\% & 75\%/60\% & 16.4 \\
        \bottomrule
        \end{tabular}
    }
    \vspace{-3mm}
    \captionof{table}{SN: ratio of matching blocks with textual/structural errors. FN: average redundant formatting commands per page.}
    \label{tab:noise_frequency}
\end{table}

\section{Additional Experimental Results}
\subsection{Experimental details}
For MinerU, we use version 0.9.2\footnote{\url{https://github.com/opendatalab/MinerU/releases/tag/magic_pdf-0.9.2-released}} by default. 
For Marker, version 0.2.17\footnote{\url{https://github.com/VikParuchuri/marker/releases/tag/v0.2.17}} is employed. 
For Nougat, we utilize its 0.1.0-base model (350M).
All prompt templates can be found in~\cref{appendix:prompt}.

For all LLMs and VLMs, we set the temperature to 0 with \texttt{do\_sample=False} by default for reproducibility.

\subsection{Sim2Real GAP}
As the questions posed by human users could have far more diversity in styles than LLM generated Q\&As.
We randomly pick 100 Q\&As and manually rewrite questions for comparison.
The performance before and after rewriting is: 27.2/23.2(GT), 20.7/18.0(MinerU), 12.8/12.9(GOT), and 23.1/20.0(Qwen2.5-VL).
Although performance degrade, the conclusions about different OCR solutions still hold, as question styles may primarily be associated with models' ability to understand instructions.

\subsection{Multimodal RAG}
We compare VisRAG with OCR-based RAG, using Qwen2-VL-7B as the generator for fair comparison.
The results are shown in~\cref{tab:visrag}.
VisRAG achieves competitive results on multimodal element-related Q\&As (e.g. table and chart), but underperforms on TXT and RO (e.g. high-resolution newspapers), exhibiting similar failure modes to Qwen2.5-VL.
\begin{table}[t]
    \centering
    \resizebox{\linewidth}{!}{
        \begin{tabular}{lcccccc}
            \toprule
            & TXT$\uparrow$ & TAB$\uparrow$ & FOR$\uparrow$ & CHA$\uparrow$ & RO$\uparrow$ & ALL$\uparrow$  \\
            \midrule
            GT          & 57.7 & 41.7 & 41.8 & 39.8 & 29.8 & 46.8 \\
            MinerU      & \underline{51.2} & 33.7 & \underline{32.1} & 10.7 & \textbf{29.8} & 37.9 \\
            Qwen2.5-VL  & \textbf{54.1} & \underline{38.9} & \textbf{34.5} & \underline{22.7} & 13.3 & \textbf{40.6} \\
            \rowcolor{gray!20} VisRAG      & 50.7 & \textbf{40.3} & 32.0 & \textbf{30.6} & \underline{15.6} & \underline{40.2} \\
            \bottomrule
        \end{tabular}
    }
    \caption{Performance of VisRAG and OCR-based RAG. We use Qwen2-VL-7B as the generator for fair comparison.}
    \label{tab:visrag}
\end{table}

\subsection{Effectiveness of robust generator}
We employ Ext2Gen-8B-R2~\cite{song2025ext2gen} and show its performance in~\cref{tab:azure_ext2gen}.
Ext2Gen-8B-R2 consistently improves performance.
Although it is based on Llama3.1-8B, its performance on Azure remains stable, reinforcing that stronger models exhibit greater robustness to formatting noise.
This further supports our conclusion that stronger models are more robust to formatting noise.
However, the performance gap between the best OCR (Azure) and GT also increases by 1.6 compared to Qwen2-7B, indicating that OCR quality becomes a bottleneck and leaves a room for improvement.
\begin{table}[t]
    \centering
    \resizebox{\linewidth}{!}{
        \begin{tabular}{lcccccccc}
        \toprule
        OCR & E.D. & TXT$\uparrow$ & TAB$\uparrow$ & FOR$\uparrow$ & CHA$\uparrow$ & RO$\uparrow$ & ALL$\uparrow$ & $\Delta$(ALL) \\
        \midrule
        \multicolumn{9}{c}{Generator: Qwen2-7B/Llama3.1-8B} \\
        \midrule
        GT         & -    & 46.7/43.1 & 31.8/37.4 & 27.6/28.4 & 31.1/34.7 & 23.7/13.7 & 36.2/35.9 & - \\
        MinerU     & 0.24 & 42.2/\ud{37.8} & 27.0/\ud{30.0} & \ud{23.5}/\ud{22.5} & 8.9/9.7   & \ud{23.0}/\tb{12.5} & 30.5/\ud{28.5} & -5.7/-7.4 \\
        Qwen2.5-VL & 0.18 & \ud{42.5}/\bf{38.6} & \ud{29.1}/\tb{33.1} & \tb{26.1}/\tb{26.1} & \ud{18.5}/\tb{19.6} & 10.9/6.7  & \ud{31.5}/\tb{30.7} & -4.7/-5.2 \\
        Azure      & 0.17 & \tb{45.5}/29.6 & \tb{30.7}/25.4 & 23.3/21.9 & \tb{19.1}/\ud{11.0} & \tb{23.5}/\ud{11.5} & \tb{33.8}/24.0 & -2.4/-11.9 \\
        \midrule
        \multicolumn{9}{c}{Generator: Ext2Gen-8B-R2} \\
        \midrule
        GT          & -    & 56.3 & 45.4 & 40.7 & 38.9 & 27.4 & 46.8 & - \\
        MinerU      & 0.24 & 49.7 & 36.4 & 30.5 & 10.8 & \ud{25.8} & 37.6 & -9.2 \\
        Qwen2.5-VL  & 0.18 & \ud{52.7} & \ud{41.2} & \tb{34.8} & \tb{24.6} & 12.9 & \ud{40.9} & -5.9 \\
        Azure       & 0.17 & \tb{55.1} & \tb{41.9} & \ud{32.4} & \ud{23.4} & \tb{26.0} & \tb{42.8} & -4.0 \\
        \bottomrule
        \end{tabular}
    }
    \caption{Experiments of Azure and Ext2Gen.}
    \label{tab:azure_ext2gen}
\end{table}

\subsection{Commercial OCR}
We evaluate Azure OCR in~\cref{tab:azure_ext2gen} and observe the following:
With powerful generators (Qwen2-7B and Ext2Gen-8b-R2), Azure yields the best performance, though there remains a gap of up to 4.0 compared to GT.
But, when using Llama3.1-8B, performance drops significantly, even worse than MinerU.
Our manual check suggests this may be due to custom formatting tags in Azure's outputs, affecting Llama3.1-8B's generation.

\label{appendix:additional_exp}
\subsection{Details in different domains}
\cref{tab:ret_domain_performance},~\cref{tab:gen_domain_performance} and ~\cref{tab:end_domain_performance} shows the performance of different OCR solution on different domains respectively.

\begin{table*}[ht]
    \centering
    \begin{tabular}{lrrrrrrr}
    \toprule
    Domain & GT & MinerU & Marker & GOT & Nougat & Qwen2.5-VL & InternVL2.5 \\
    \midrule
    Law & 81.2 & 71.0 & 77.1 & 62.1 & 69.0 & 76.4 & 69.6 \\
    Finance & 59.7 & 36.4 & 45.0 & 30.4 & 25.8 & 47.9 & 47.1 \\
    Textbook & 73.2 & 43.8 & 49.6 & 48.8 & 37.1 & 58.3 & 55.0 \\
    Manual & 79.1 & 60.4 & 68.6 & 58.9 & 47.8 & 71.3 & 70.2 \\
    Newspaper & 40.5 & 31.3 & 34.0 & 12.4 & 10.6 & 27.7 & 18.4 \\
    Academic & 75.1 & 50.3 & 55.2 & 50.2 & 45.0 & 61.1 & 57.1 \\
    Administration & 82.2 & 59.4 & 68.3 & 57.7 & 52.7 & 73.1 & 73.8 \\
    All & 70.0 & 50.1 & 56.6 & 45.4 & 40.8 & 59.2 & 55.8 \\
    \bottomrule
    \end{tabular}
    \caption{Retrieval performance across different domains.}
    \label{tab:ret_domain_performance}
\end{table*}

\begin{table*}[ht]
    \centering
    \begin{tabular}{lrrrrrrr}
    \toprule
    Domain & GT & MinerU & Marker & GOT & Nougat & Qwen2.5-VL & InternVL2.5 \\
    \midrule
    Law & 56.9 & 53.4 & 54.4 & 43.3 & 48.8 & 53.9 & 50.9 \\
    Finance & 43.1 & 30.1 & 29.5 & 19.7 & 17.7 & 35.9 & 36.8 \\
    Textbook & 37.6 & 25.9 & 28.2 & 24.8 & 16.8 & 29.1 & 29.1 \\
    Manual & 50.2 & 45.3 & 46.1 & 41.3 & 34.3 & 48.7 & 47.7 \\
    Newspaper & 35.0 & 33.7 & 31.6 & 9.5 & 8.4 & 19.6 & 11.7 \\
    Academic & 38.3 & 29.5 & 27.9 & 25.3 & 24.8 & 33.2 & 31.3 \\
    Administration & 46.4 & 35.7 & 37.7 & 32.2 & 29.2 & 42.7 & 42.9 \\
    All & 43.9 & 36.1 & 36.3 & 27.8 & 25.5 & 37.5 & 35.8 \\
    \bottomrule
    \end{tabular}
    \caption{Generation performance across different domains.}
    \label{tab:gen_domain_performance}
\end{table*}

\begin{table*}[ht]
    \centering
    \begin{tabular}{lrrrrrrr}
    \toprule
    Domain & GT & MinerU & Marker & GOT & Nougat & Qwen2.5-VL & InternVL2.5 \\
    \midrule
    Law & 49.6 & 48.1 & 48.1 & 41.1 & 43.9 & 47.2 & 44.9 \\
    Finance & 27.2 & 19.4 & 20.1 & 15.1 & 13.1 & 22.9 & 22.8 \\
    Textbook & 30.5 & 20.9 & 22.5 & 21.0 & 15.7 & 23.8 & 23.5 \\
    Manual & 44.4 & 38.1 & 39.8 & 36.0 & 30.7 & 42.3 & 41.6 \\
    News & 29.0 & 25.6 & 24.7 & 8.3 & 5.6 & 17.4 & 11.0 \\
    Academic & 31.9 & 25.6 & 24.1 & 22.8 & 21.2 & 27.6 & 26.4 \\
    Administration & 41.0 & 30.9 & 32.7 & 29.2 & 26.6 & 37.3 & 37.5 \\
    All & 36.1 & 29.5 & 30.0 & 24.6 & 22.2 & 31.1 & 29.6 \\
    \bottomrule
    \end{tabular}
    \caption{Overall performance across different domains.}
    \label{tab:end_domain_performance}
\end{table*}


\begin{table*}[t]
    \centering
    \begin{tabular}{p{0.9\linewidth}}
    \toprule
    \sethlcolor{yellow}\hl{\textbf{System:}} \\
    \quad You are an AI specialized in generating QAs from documents. Your mission is to analyze the document, follow the instructions, and generate RAG-style question-answer pairs based on the document. \\
    \\
    \quad RAG-style refers to a question that needs to be answered by retrieving relevant context from an external document based on the question, so the question MUST obey the following criteria: \\
    \quad 1. Question should represent a plausible inquiry that a person (who has not seen the page) might ask about the information uniquely presented on this page. The questions should not reference this specific page directly (by page number, pointing to a specific paragraph or figure, and never refer to the document using phrases like 'in the document'), nor should they quote the text verbatim. They should use natural language reflecting how someone might inquire about the page’s content without direct access. \\
    \quad 2. Question must contain all information and context/background necessary to answer without the document. Do not include phrases like "according to the document" in the question. \\
    \quad 3. Question must not contain any ambiguous references, such as 'he', 'she', 'it', 'the report', 'the paper', and 'the document'. You MUST use their complete names.
    \\
    \sethlcolor{yellow}\hl{\textbf{User:}}\\
    \quad Your task is to generate several RAG-style question-answer pairs with different levels of difficulty and evidence sources. \textcolor{blue}{\{detailed\_task\_description\}}.\\
    \\
    You MUST obey the following criteria: \\
    \quad - The question MUST be detailed and be based explicitly on information in the document. \\
    \quad - The question MUST include at least one entity. \\
    \quad - The context sentence the question is based on MUST include the name of the entity. For example, an unacceptable context is "He won a bronze medal in the 4 × 100 m relay". An acceptable context is "Nils Sandström was a Swedish sprinter who competed at the 1920 Summer Olympics." \\
    \quad - The answer form should be as diverse as possible, including [Yes/No, Numeric, String, List]. \\
    \quad - \textcolor{blue}{\{additional\_task\_criteria\}}\\
    \\
    \quad If there are no possible questions that meet these criteria, return 'None' as the question. Output the question in JSON format. \\
    \\
    \quad \textcolor{blue}{\{qa\_examples\}} \\
    \\
    \quad <document>\textcolor{blue}{\{document\}}</document>\\
    
    \bottomrule
    \end{tabular}
    \caption{Q\&A Generation Prompt}
    \label{tab:qa_gen_prompt}
\end{table*}

\begin{table*}[h!]
    \centering
    \begin{tabular}{p{0.9\linewidth}}
    \toprule
    \sethlcolor{yellow}\hl{\textbf{Structure data task:}} \\
    \quad In the given documents, the chart elements are all enclosed within <chart> </chart> tags and illustrated in LaTeX table format. Pay attention to the difference between them and tabular data, as tabular data is not enclosed by <chart> </chart> tags. \sethlcolor{yellow}\hl{\# This paragraph is only used for chart data.} \\
    \quad In order to generate this type of question-answer pairs, first, you need to read the given document, identify the table/formula/chart elements within it, and use them as the evidence context. The evidence context can be a single paragraph for single-hop questions, or several related paragraphs for generating multi-hop questions that require reasoning. After that, you need to generate questions and corresponding answers based on them. \\
    \\
    \hline
    \sethlcolor{yellow}\hl{\textbf{Reading order task:}} \\
    \quad Your task is to generate RAG-style question-answer pairs from the given two documents. \\
    In order to generate this type of question-answer pairs, first, you need to read the given two documents (A, B), identify the text sharing the same entities, and design a question-answer pair based on the contents of both documents A and B. If it is based on the message of document A or document B alone, it cannot be answered.\\
    \\
    \hline
    \sethlcolor{yellow}\hl{\textbf{Understanding task:}}\\
    \quad You should generate question-answering pairs that require the responser to extract information from documents. The answer should be able to find directly in the documents without any reasoning.\\
    \\
    \hline
    \sethlcolor{yellow}\hl{\textbf{Reasoning task:}}\\
    \quad You should generate question-answering pairs that require responser to reason before answering, such as calculations, comparisons, finding the maximum and minimum, or integration information from different parts of the documents. The answer should not be able to be found directly in the documents. \\
    \bottomrule
    \end{tabular}
    \caption{Detailed description used to generate Q\&A pairs for different tasks.}
    \label{tab:task_prompt}
\end{table*}

\begin{table*}[h!]
    \centering
    \begin{tabular}{p{0.9\linewidth}}
    \toprule
    \sethlcolor{yellow}\hl{\textbf{Multi-page Q\&A from single-page question:}}\\
    \quad Your mission is to generate RAG-style combined questions from two questions that have the same entity. \\
    \\
    \quad When generating a combined question, there are some criteria you should follow: \\
    \quad - The answer to the combined question should be the same as the answer2. \\
    \quad - It must combine the answer1 to question1 to answer the combined questions. This means that, to answer the combined question, a responder must first deduce the part of the combined question that refers to the answer1, and then proceed to answer the combined question based on that answer. \\
    \quad - You cannot include the answer to question 1 in the combined question. \\
    \textcolor{blue}{\{combined\_qa\_examples\}} \\
    \quad Based on the above 3 examples, provide a combined question for the following case. If you find it is hard to create such a combined question, output None as the answer. Enclose the combined question within <answer> </answer>: \\
    \quad question1: \textcolor{blue}{\{q1\}} \\
    \quad answer1: \textcolor{blue}{\{a1\}} \\
    \quad question2: \textcolor{blue}{\{q2\}} \\
    \quad answer2: \textcolor{blue}{\{a2\}} \\
    \\
    \hline
    \sethlcolor{yellow}\hl{\textbf{Multi-page Q\&A from different paragraphs:}}\\
    \quad Your task is to generate RAG-style question-answer pairs from the given two documents and entity names. The entity names appear in both documents, and you need to use them as a bridge to generate the RAG-style question-answer pairs that need to be answered by combining information from both documents. \\
    \quad To generate the question-answer pairs, first, you need to read the given two documents (A, B) and the entity names, find paragraphs related to them, use the paragraphs as evidence context, and design a question-answer pair based on the evidence context from the two documents. \\
    \bottomrule
    \end{tabular}
    \caption{Detailed description used to generate multi-page Q\&A pairs from both single-page questions and different paragraphs sharing same entities.}
    \label{tab:mp_prompt}
\end{table*}

\begin{table*}[t]
    \centering
    \begin{tabular}{p{0.9\linewidth}}
    \toprule
    \sethlcolor{yellow}\hl{\textbf{System:}} \\
    \quad You are an expert, you have been provided with a question and documents retrieved based on that question. Your task is to search the content and answer these questions using the retrieved information. \\

    \quad You **MUST** answer the questions briefly with one or two words or very short sentences, devoid of additional elaborations. \\
    
    \quad Write the answers within <response></response>. \\
    \\
    \sethlcolor{yellow}\hl{\textbf{User:}}\\
    \quad Question: \textcolor{blue}{\{question\}}\\
    \quad Retrieved Documents: \textcolor{blue}{\{retrieved\_documents\}}\\
    
    \bottomrule
    \end{tabular}
    \caption{LLMs prompt for RAG generation}
    \label{tab:llm_prompt}
\end{table*}

\begin{table*}[h!]
    \centering
    \begin{tabular}{p{0.9\linewidth}}
    \toprule
    \sethlcolor{yellow}\hl{\textbf{Simple Prompt:}} \\
    \quad Please do OCR on the image and give all the text content in markdown format. The formulas should be wrapped in \$\$. The table and charts should be parsed in LaTeX format. Only output the OCR results without any extra explanations or comments.\\
    \\
    \bottomrule
    \end{tabular}
    \caption{Simple prompt for VLMs OCR}
    \label{tab:ocr_prompt}
\end{table*}

\begin{table*}[h!]
    \centering
    \begin{tabular}{p{0.9\linewidth}}
    \toprule
    \sethlcolor{yellow}\hl{\textbf{Detailed Prompt:}}\\
    \quad You are a powerful OCR assistant tasked with converting PDF images to the Markdown format. You MUST obey the following criteria:
    \\
    \quad 1. Plain text processing:\\
    \quad - Accurately recognize all text content in the PDF image without guessing or inferring. \\
    \quad - Precisely recognize all text in the PDF image without making assumptions in the Markdown format.\\
    \quad - Maintain the original document structure, including headings, paragraphs, lists, etc. \\

    \quad 2. Formula Processing:\\
    \quad - Convert all formulas to LaTeX.\\
    \quad - Enclose inline formulas with \$ \$. For example: This is an inline formula \$ E = mc\^2 \$.\\
    \quad - Enclose block formulas with \$\$ \$\$. For example: \$\$ \textbackslash frac\{-b \textbackslash pm \textbackslash sqrt\{b\^2 - 4ac\}\}\{2a\} \$\$.\\

    \quad 3. Table Processing:\\
    \quad - Convert all tables to LaTeX format.\\
    \quad - Enclose the tabular data with \textbackslash begin\{table\} \textbackslash end\{table\}.

    \quad 3. Chart Processing:\\
    \quad - Convert all Charts to LaTeX format.\\
    \quad - Enclose the chart data in tabular with \textbackslash begin\{table\} \textbackslash end\{table\}.

    \quad 4. Figure Handling:\\
    \quad - Ignore figures from the PDF image; do not describe or convert images.\\

    \quad 5. Output Format:\\
    \quad - Ensure the Markdown output has a clear structure with appropriate line breaks.\\
    \quad - Maintain the original layout and format as closely as possible.\\

    \quad Please strictly follow these guidelines to ensure accuracy and consistency in the conversion. Your task is to accurately convert the content of the PDF image using these format requirements without adding any extra explanations or comments.\\
    
    \bottomrule
    \end{tabular}
    \caption{Complex prompt for VLMs OCR}
    \label{tab:ocr_prompt_comp}
\end{table*}

\begin{table*}[t]
    \centering
    \begin{tabular}{p{0.9\linewidth}}
    \toprule
    \sethlcolor{yellow}\hl{\textbf{System:}} \\
    \quad You are an AI specialized in document question-answering verification. Your mission is to analyze the given question-answering pairs and follow the instructions. Your response must be true and accurate, and no additional content should be output. \\
    \\
    1. Question type check \\
    \quad Dose the question match the task description: \textcolor{blue}{\{detailed\_task\_description\}} \\
    \quad Make sure the question meets the required task context. \\
    2. Evidence relevance Check \\
    \quad Dose the provided evidence context relate to the question provided? Does the answer accurately reflect the information in the evidence context? Ensure the question is formulated based on information explicitly stated. The question should not introduce concepts unrelated to the document's content. \\
    3. Clarity and Precision \\
    \quad Is the question clear and unambiguous? And is the answer concise and precise? Ensure the language is straightforward and easily understandable, and avoid complex phrasing that may confuse the reader.
    \quad The intention of the question and answer pair must be clear and direct, avoiding verbosity and unnecessary detail.
    \quad Ensure the answer fully addresses the question without omitting crucial information.\\
    \quad \textcolor{blue}{\{qas\}} \\
    \\
    
    \bottomrule
    \end{tabular}
    \caption{Q\&A Verification Prompt}
    \label{tab:qa_verification_prompt}
\end{table*}

\section{Case Study}
\begin{figure*}[t]
    \centering
    \includegraphics[width=0.9\linewidth]{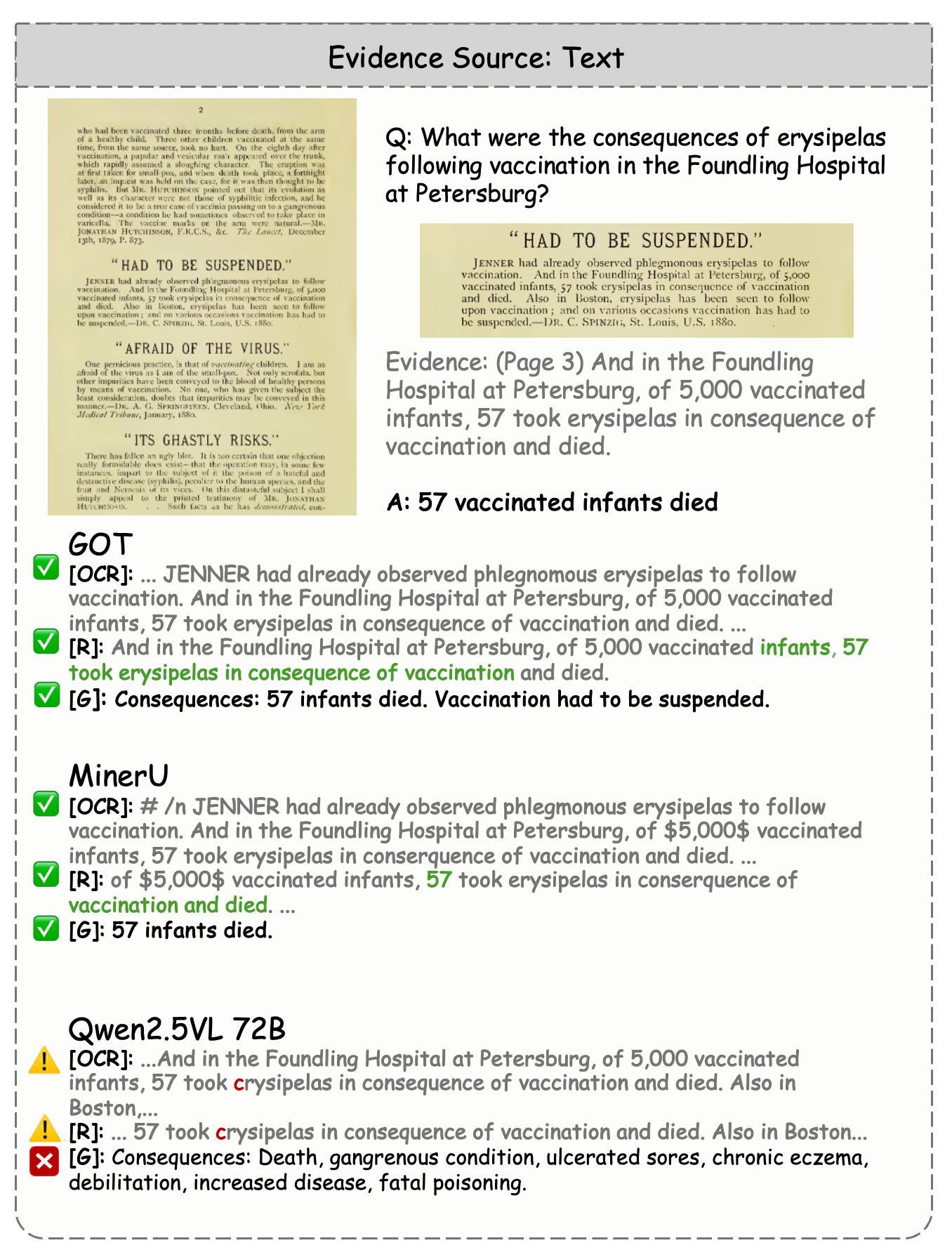}
    \caption{A case using text as the evidence source on a distorted academic document.}
    \label{fig:case_1}
\end{figure*}

\begin{figure*}[t]
    \centering
    \includegraphics[width=0.9\linewidth]{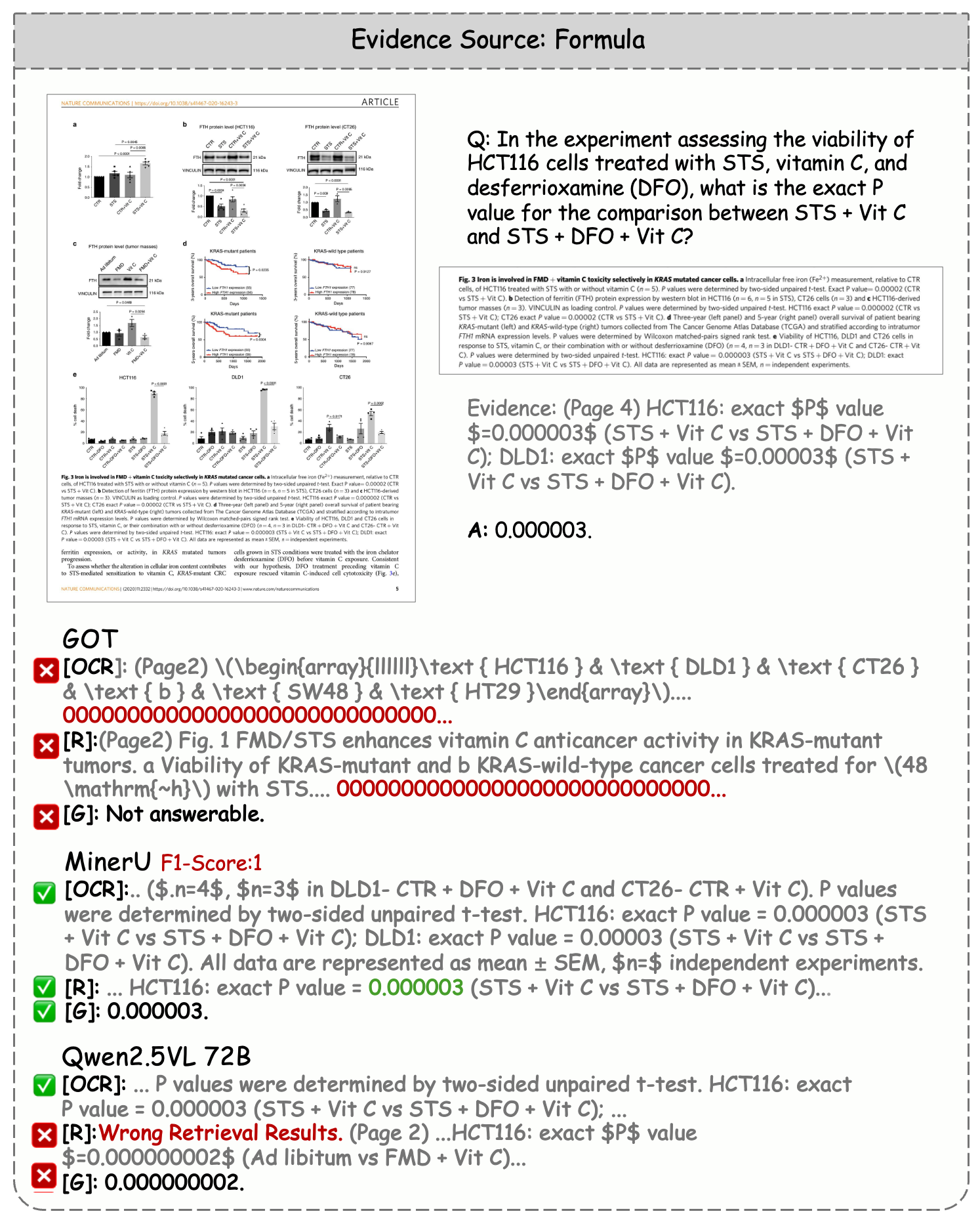}
    \caption{A case using formula as the evidence source on a scanned academic document.}
    \label{fig:case_2}
\end{figure*}

\begin{figure*}[t]
    \centering
    \includegraphics[width=0.9\linewidth]{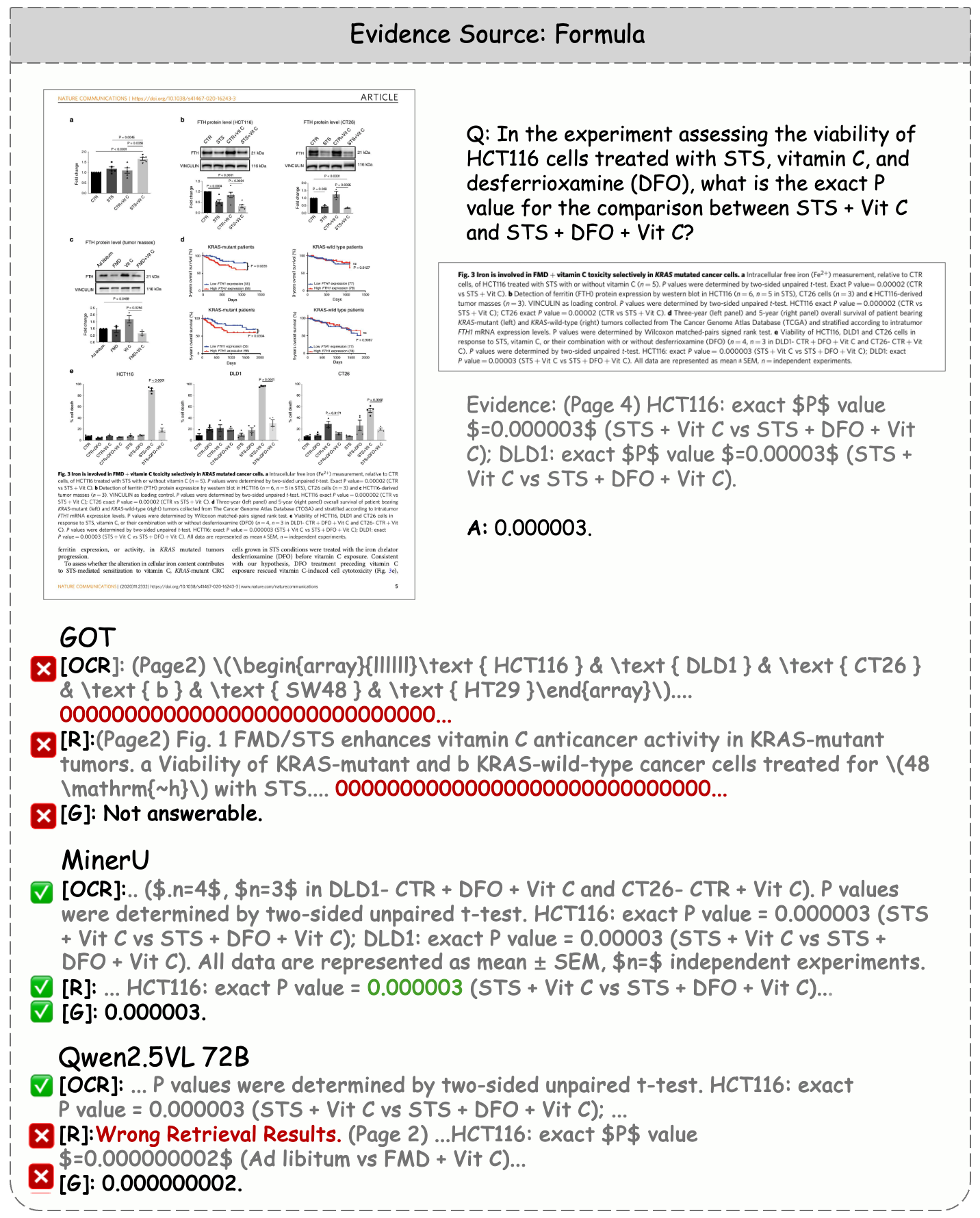}
    \caption{A case using text in multi-pages as the evidence source on an academic document.}
    \label{fig:case_3}
\end{figure*}

\begin{figure*}[t]
    \centering
    \includegraphics[width=0.87\linewidth]{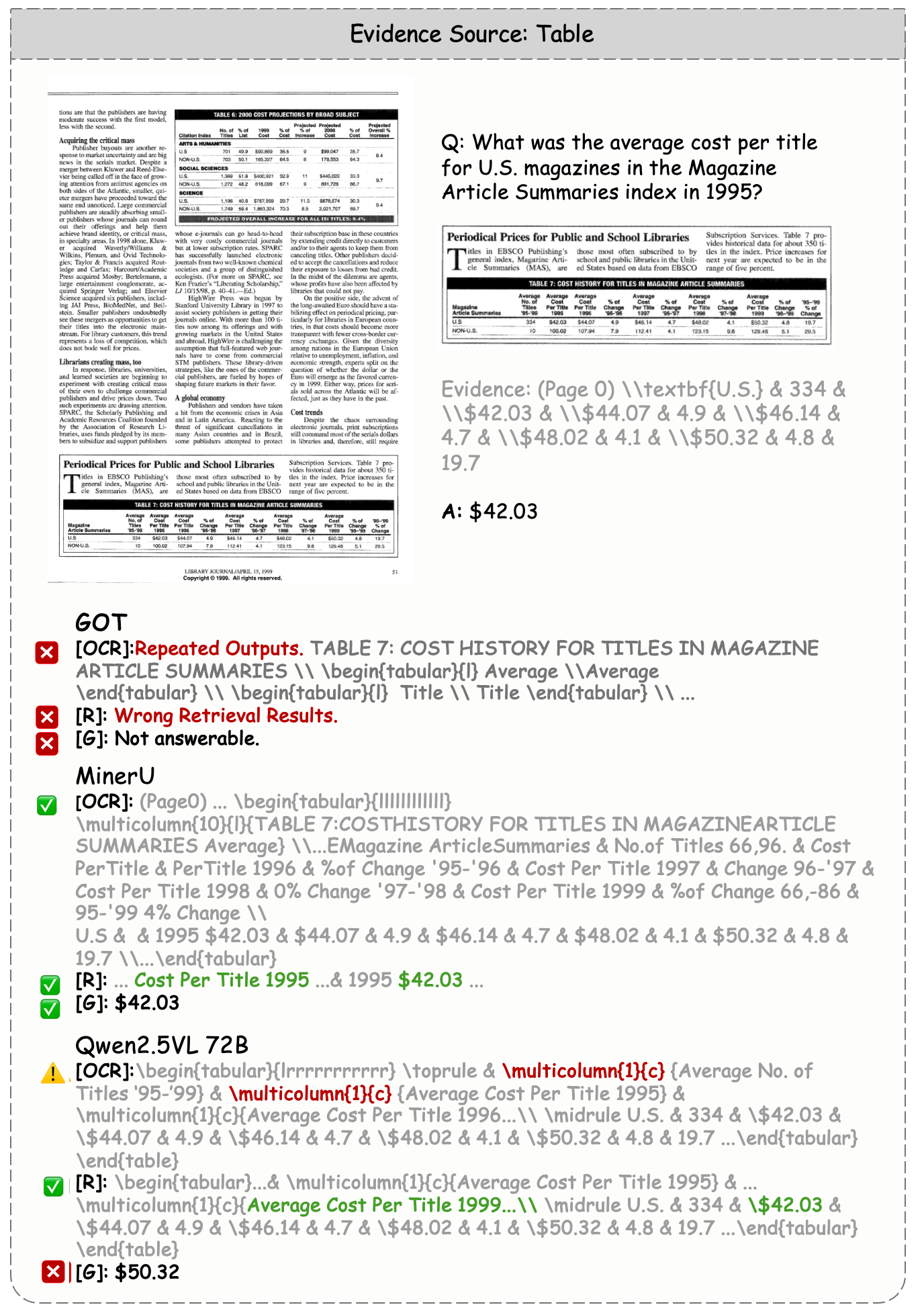}
    \caption{A case using formula as the evidence source on a scanned textbook.}
    \label{fig:case_4}
\end{figure*}

\begin{figure*}[t]
    \centering
    \includegraphics[width = \linewidth]{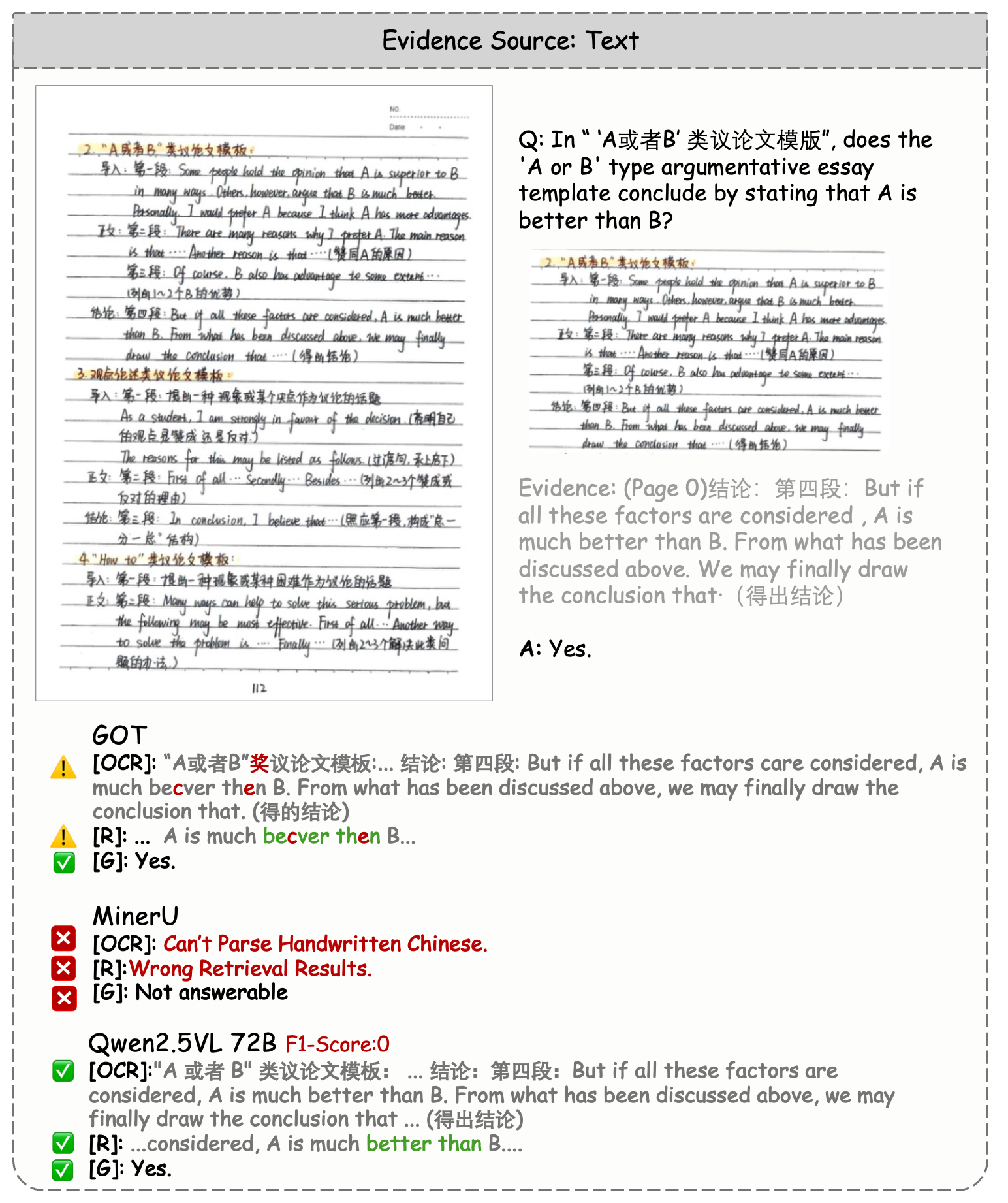}
    \caption{A case using text as the evidence source on a handwritten textbook.}
    \label{fig:case_5}
\end{figure*}

\begin{figure*}[t]
    \centering
    \includegraphics[width = \linewidth]{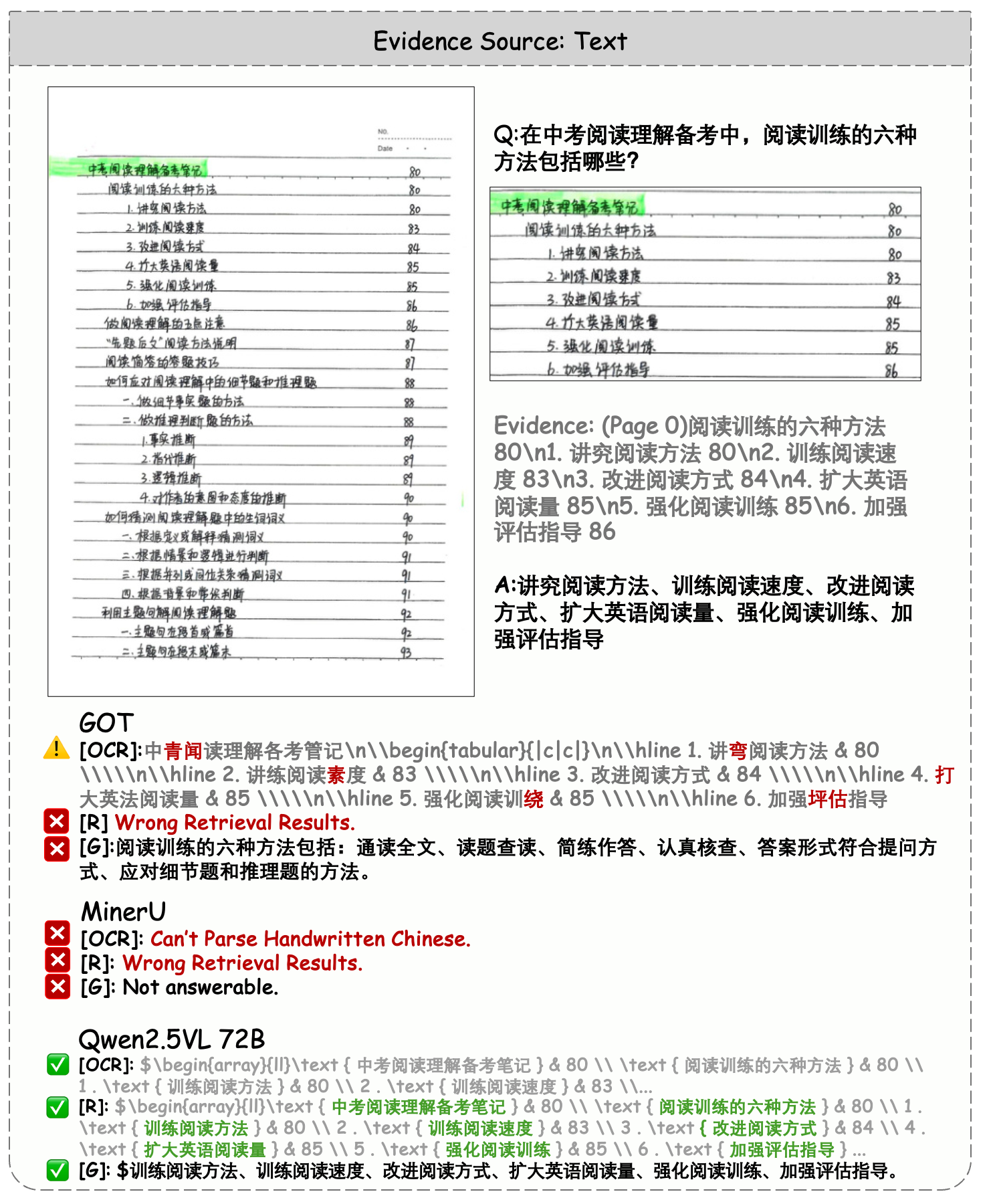}
    \caption{A case using text as the evidence source on a handwritten textbook.}
    \label{fig:case_6}
\end{figure*}

\begin{figure*}[t]
    \centering
    \includegraphics[width = \linewidth]{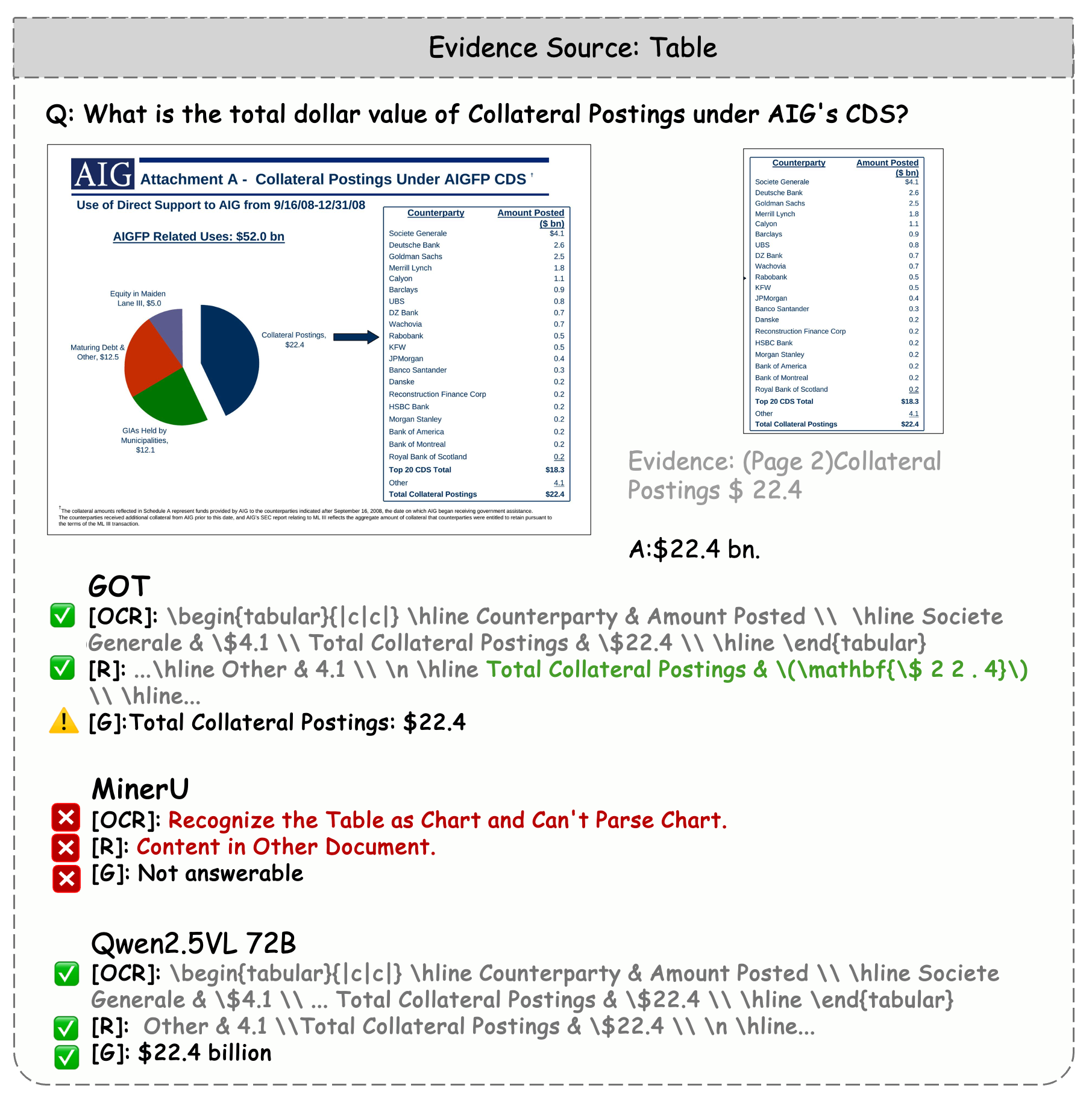}
    \caption{A case using table as the evidence source on a financial report.}
    \label{fig:case_7}
\end{figure*}

\begin{figure*}[t]
    \centering
    \includegraphics[width = \linewidth]{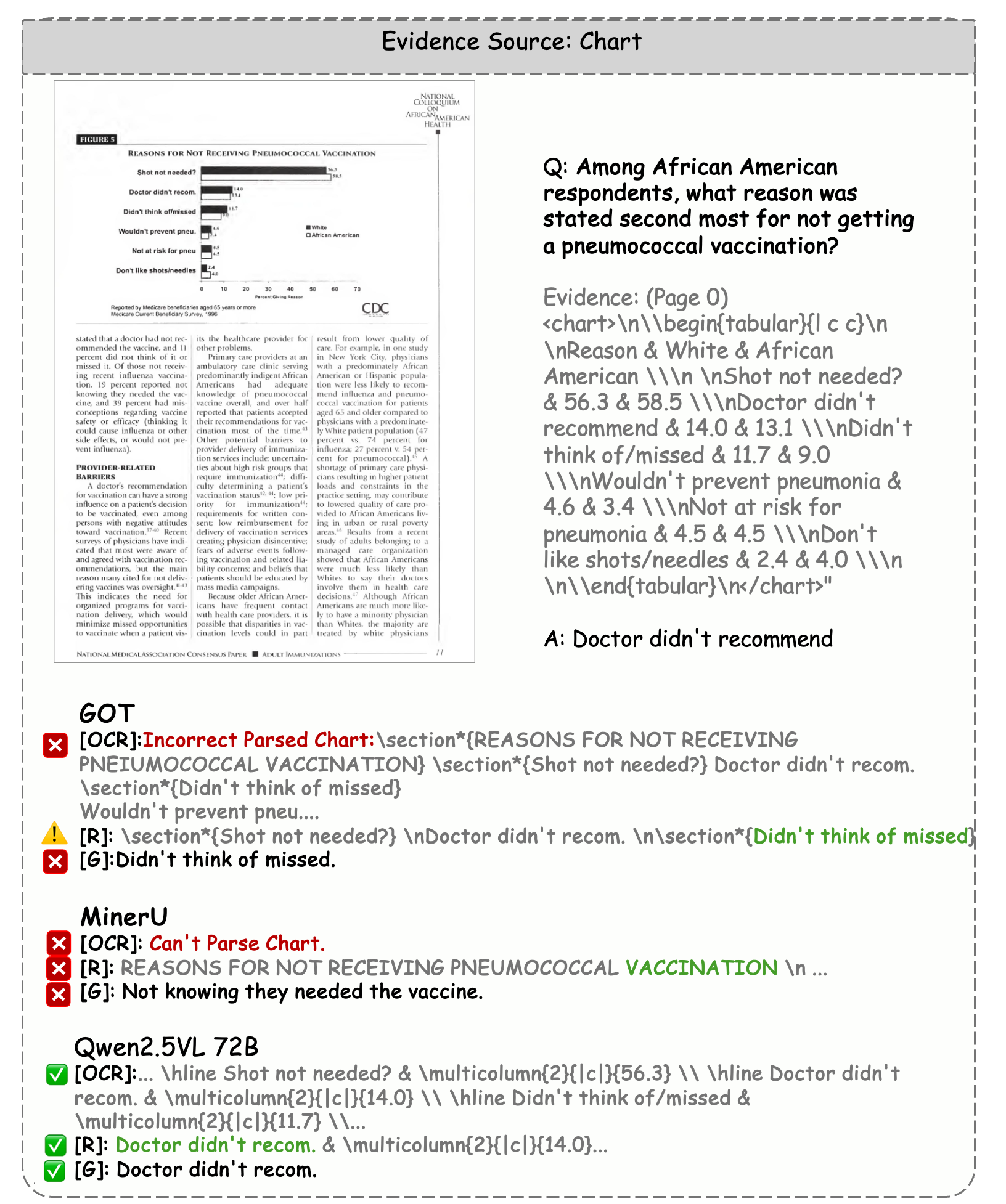}
    \caption{A case using table as the evidence source on a scanned academic paper.}
    \label{fig:case_8}
\end{figure*}

\begin{figure*}[t]
    \centering
    \includegraphics[width = 0.9\linewidth]{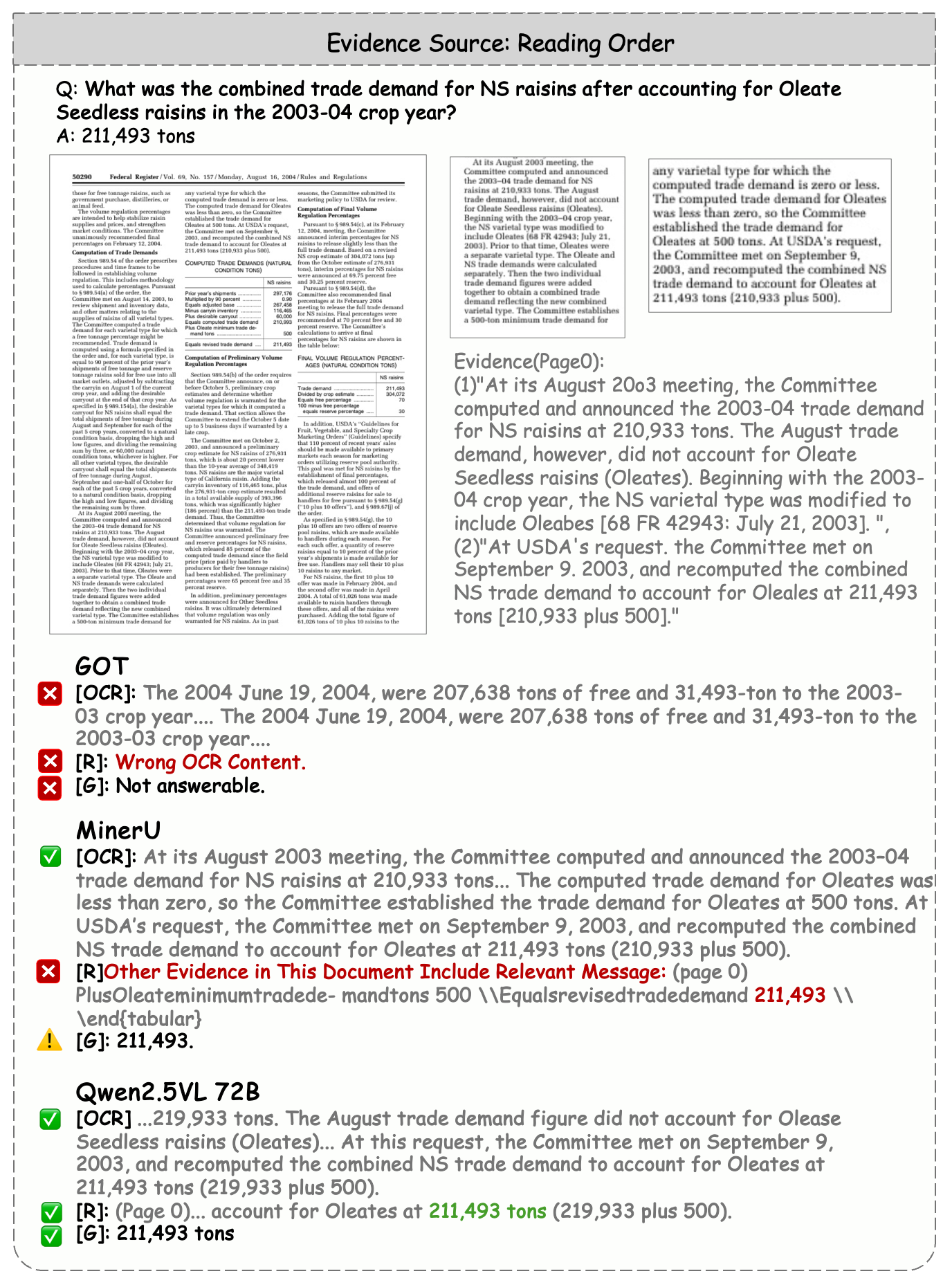}
    \caption{A case using text with reading order as the evidence source on a scanned newspaper.}
    \label{fig:case_9}
\end{figure*}

\begin{figure*}[t]
    \centering
    \includegraphics[width = \linewidth]{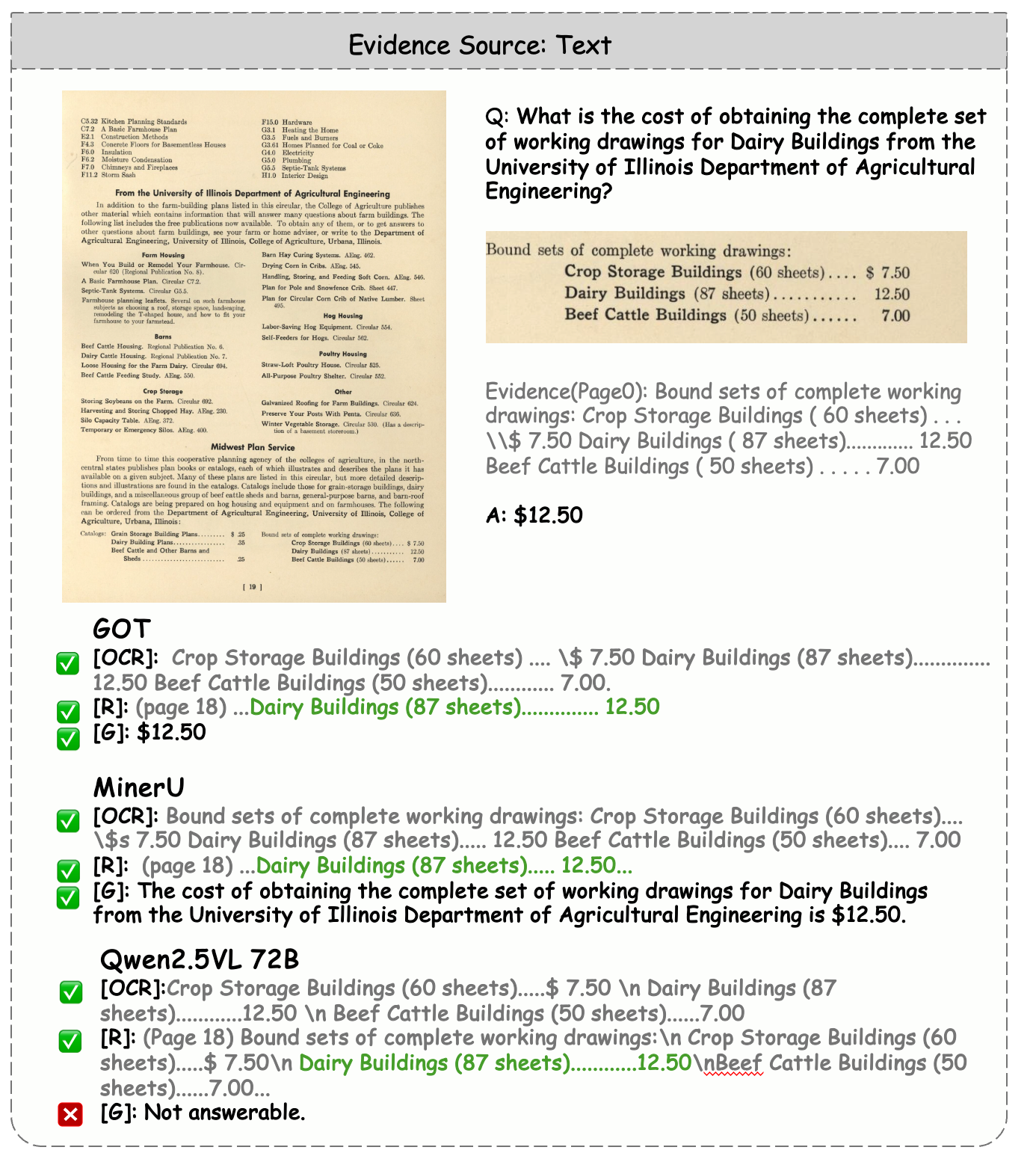}
    \caption{A case using text as the evidence source on a distortion manual.}
    \label{fig:case_10}
\end{figure*}

\cref{fig:case_1} to ~\cref{fig:case_10} show some cases of GOT, MinerU, and Qwen2.5VL-72B on OHRbench. For each case, we indicate the evidence source and answer, giving the OCR result of different models and the responses at the retrieval and generation stages.

\end{document}